\documentclass[lettersize,journal]{IEEEtran}
\usepackage{amsmath,amssymb,amsfonts}
\usepackage{amsthm}
\usepackage{multirow}
\usepackage{tabularx}
\usepackage{algorithmic}
\usepackage{algorithm}
\usepackage{caption}
\usepackage{array}
\usepackage[caption=false,font=normalsize,labelfont=sf,textfont=sf]{subfig}
\usepackage{textcomp}
\usepackage{stfloats}
\usepackage{url}
\usepackage{verbatim}
\usepackage{graphicx}
\newtheorem{definition}{Definition}
\graphicspath{{./image/}}
\usepackage{color}

\makeatletter
\newenvironment{breakablealgorithm}
{% \begin{breakablealgorithm}
\begin{center} %
\refstepcounter{algorithm}% New algorithm
\hrule height.8pt depth0pt \kern2pt% \@fs@pre for \@fs@ruled
\renewcommand{\caption}[2][\relax]{% Make a new \caption
{\raggedright\textbf{\ALG@name~\thealgorithm} ##2\par}

\ifx\relax##1\relax

\addcontentsline{loa}{algorithm}{\protect\numberline{\thealgorithm}##2}%

\else

\addcontentsline{loa}{algorithm}{\protect\numberline{\thealgorithm}##1}%

\fi

\kern2pt\hrule\kern2pt

}

}{% \end{breakablealgorithm}

\kern2pt\hrule\relax% \@fs@post for \@fs@ruled

\end{center} %

}

\makeatother

\def\BibTeX{{\rm B\kern-.05em{\sc i\kern-.025em b}\kern-.08em
    T\kern-.1667em\lower.7ex\hbox{E}\kern-.125emX}}
\usepackage{balance}

\begin{document}
\title{Guiding Multi-agent Multi-task Reinforcement Learning by a Hierarchical Framework with Logical Reward Shaping}
%\title{A Multiagent Reinforcement Learning algorithm for Multi-task solving based on a Hierarchical Framework with Logical Reward Shaping}

\author{Chanjuan Liu, \IEEEmembership{Member, IEEE}, Jinmiao Cong, Bingcai Chen, Yaochu Jin, \IEEEmembership{Fellow, IEEE} and  Enqiang Zhu*
\thanks{This work was supported in part by the National Natural Science Foundation of China (No.2172072), in part by Natural Science Foundation of Liaoning Province of China under Grant 2021-MS-114, and in part by Dalian Youth Star of Science and Technology 2020RQ063.}
\thanks{Chanjuan Liu is with the School of Computer Science and Technology, Dalian University of Technology, Dalian 116024, China (e-mail: chanjuanliu@dlut.edu.cn). }
\thanks{Jinmiao Cong is with the School of Computer Science and Technology, Dalian University of Technology, Dalian 116024, China.}
\thanks{Bingcai Chen is with the School of Computer Science and Technology, Dalian University of Technology, Dalian 116024, China.}
\thanks {Yaochu Jin is with the School of Engineering, Westlake University, Hangzhou 310024, China(e-mail: jinyaochu@westlake.edu.cn).}
\thanks{Enqiang Zhu is with 
the Institute of Computing Technology, Guangzhou University, Guangzhou 510006, China (e-mail: zhuenqiang@gzhu.edu.cn).}}

\markboth{Journal of \LaTeX\ Class Files,~Vol.~18, No.~9, September~2020}%
{How to Use the IEEEtran \LaTeX \ Templates}

\maketitle

\begin{abstract}

Multi-agent hierarchical reinforcement learning (MAHRL) has been studied as an effective means to solve intelligent decision problems in complex and large-scale environments. However, most current MAHRL algorithms follow the traditional way of using reward functions in reinforcement learning, which limits their use to a single task.  This study aims to design a multi-agent cooperative algorithm with logic reward shaping (LRS), which uses a more flexible way of setting the rewards, allowing for the effective completion of multi-tasks.  LRS uses Linear Temporal Logic (LTL) to express the internal logic relation of subtasks within a complex task. Then, it evaluates whether the subformulae of the LTL expressions are satisfied based on a designed reward structure. This helps agents to learn to effectively complete tasks by adhering to the LTL expressions, thus enhancing the interpretability and credibility of their decisions. To enhance coordination and cooperation among multiple agents, a value iteration technique is designed to evaluate the actions taken by each agent. Based on this evaluation, a reward function is shaped for coordination, which enables each agent to evaluate its status and complete the remaining subtasks through experiential learning. Experiments have been conducted on various types of tasks in the Minecraft-like environment. The results demonstrate that the proposed algorithm can improve the performance of multi-agents when learning to complete multi-tasks.

\end{abstract}


\section{Introduction}
\label{sec:introduction}

Deep reinforcement learning (DRL)  has shown remarkable success in solving decision-making problems that surpass human-level performance, such as the Atari game \cite{kumar2021artificial}, chess confrontation \cite{silver2018general,8248673}, and real-time strategy game (RTS) \cite{jaderberg2019human}. However, as the environments become increasingly complex, some limitations (such as low learning efficiency and quality) may appear in single-agent DRL systems. To address this, there is an urgent need for multi-agent DRL \cite{gronauer2022multi}, where multiple agents can solve complex tasks through collaboration \cite{DBLP:conf/aaai/FuQYPW22}. However, multi-agent learning \cite{9043893} for complex tasks suffers from an exponential growth of the action and state spaces, which is known as the curse of dimensionality \cite{du2021survey}. To overcome this, hierarchical reinforcement learning (HRL) \cite{wu2019hierarchical} has been introduced into multi-agent DRL, giving rise to multi-agent hierarchical reinforcement learning (MAHRL) \cite{DBLP:journals/eswa/ZhouSTO21,han2019multi}.

\subsection{The Challenges}
 
Most existing MAHRL algorithms follow the traditional way of setting the reward functions, which is not appropriate when multiple tasks need to be completed in complex environments. For instance, in the Minecraft environment, in order to complete the task of making bows and arrows, agents have to find wood to make the body of bows and arrows, spider silk to make bowstrings, as well as feathers to make arrow fletchings. To learn the strategies for completing the task, an appropriate reward is needed for the agent. However, designing a reward function for one task is challenging and difficult to generalize for other tasks \cite{10130298}. Moreover, in the task of making bows and arrows, if an agent only finds some of the required materials, it can not get a reward; thus, it is challenging for agents to learn how to complete the remaining tasks. In MAHRL, the decision of each agent is typically treated as a black box, making it difficult to understand the logic behind this decision, leading to the untrustworthiness of the system. Hence, it is essential to develop a general and effective way of shaping rewards with a description of the internal logic of the tasks, which helps the agents easily understand the progress of the task and make reasonable decisions.

\subsection{Our Contributions}

This work explores a flexible approach to setting rewards, called logic reward shaping (LRS), for multi-task learning. LRS uses the Linear Temporal Logic (LTL) \cite{LIU2022110254, DBLP:journals/tac/CaiPLK21, chiari2020linear, yang2021reinforcement} to represent environmental tasks, making use of its precise semantics and compact syntax to clearly show the internal logical construction of the tasks and provide guidance for the agents. A reward structure is appropriately defined to give rewards, based on whether LTL expressions are satisfied or not. To promote strategy learning, a technique of value iteration is used to evaluate the actions taken by each agent; after that, a  reward shaping mechanism is utilized to shape a reward function, which can accelerate the learning and coordination between agents.

The advantage of the LRS mechanism lies in the formalization provided by LTL to specify the constraints of tasks, ensuring that the agent's decisions meet the specified requirements. Through the feedback of rewards, the agent gradually adjusts its strategy to meet the logical specifications defined in LTL. Consequently, the agent can execute tasks more reliably by adhering to the prescribed logical requirements, thus enhancing the credibility of decisions.

% Based on the LRS mechanism, \textcolor{blue}{the agent is capable of enhanced comprehension and adherence to the logical requisites of the task, facilitating more effective handling of the task's logical intricacies. This ensures that decision-making aligns with logical rules, consequently elevating the credibility of decision-making in complex environments.}

% \textcolor{blue}{The advantage of this mechanism lies in the formalization provided by LTL to specify the expected behavior during task execution, ensuring that the agent's decisions conform to the specified requirements. Through reinforcement learning's iterative trial-and-error learning process, the agent gradually adjusts its strategy to meet the logical specifications defined in LTL. Consequently, the agent can execute tasks more reliably, minimizing decisions that deviate from logical specifications and enhancing the trustworthiness and robustness of decision-making. This integration helps ensure that the agent better adheres to the prescribed logical requirements during task execution, thereby bolstering the credibility of decision-making.}

Based on LRS, we propose a multi-agent hierarchical reinforcement learning algorithm, dubbed Multi-agent Hierarchy via Logic Reward Shaping (MHLRS). In MHLRS, the agents aim to achieve joint tasks, but each maintains their own individual structures. Each agent has its own meta-controller, which learns sub-goal strategies based on the state of the environment. The experiments on different scenarios show that the proposed MHLRS enhances the cooperative performance of multi-agents in completing multi-tasks.

\subsection{Related Work}

Learning coordination in multi-agent systems is a challenging task due to increased complexity and the involvement of multiple agents. As a result, many methods and ideas have been proposed to address this issue  \cite{tan1997multi, yuan2022multi}.  Kumar et al. \cite{kumar2017federated} used a master-slave architecture to solve the coordination problem between the agents. A higher-level controller guides the information exchange between decentralized agents. Based on the guidance of the controller, each agent communicates with another agent in each time step, which allows for the exploration of distributed strategies. However, the scalability of this method remains to be improved, since information exchange between agents that are far away becomes more difficult as the number of agents increases. Budhitama et al. \cite{DBLP:journals/eswa/ZhouSTO21} also adopted a similar structure that included the commander agent and unit agent models. The commander makes decisions based on environmental states, and then the units execute those decisions. However, the commander's global decisions might need to be more suitable for some units. In the proposed MHLRS, each agent has its own meta-controller that proposes suitable sub-goal strategies according to the state of the environment and other agents.

Constructed with propositions on environmental states, logical connectors, and temporal operators, LTL  \cite{DBLP:conf/allerton/SadraddiniB15, DBLP:conf/aips/GiacomoIFP19, wen2021probably} can naturally represent the tasks in reinforcement learning. Some studies on using LTL for reinforcement learning \cite{DBLP:conf/iros/KuoKB20, cai2021reinforcement, DBLP:conf/iclr/HahnSKRF21} have been reported, where different methods have been employed to guide the RL agent to complete various tasks. Toro Icarte et al. \cite{DBLP:conf/atal/IcarteKVM18} used the co-safe LTL expression to solve agents' multi-task learning problems and introduced the extension of Q-learning, viz., LTL Progression Off-Policy Learning (LPOPL). To reduce the cost of learning LTL semantics, Vaczipoor et al. \cite{DBLP:conf/icml/VaezipoorLIM21} introduced an environment-independent LTL pre-training scheme. They utilized a neural network to encode the LTL formulae so that RL agents can learn strategies with task conditions. However, these methods are proposed for single-agent systems rather than multi-agent systems. G. Leon et al. \cite{leon2020extended} extended LTL from a single-agent framework to a multi-agent framework, and proposed two MARL algorithms that are highly relevant to our work. Nevertheless, traditional Q-learning and DQN frameworks are used, which makes it difficult for agents to explore stable collaborative strategies in dynamic environments. To address this issue, a hierarchical structure is introduced in this work to enable more flexible strategy exploration, accelerate the learning process, and enable agents to adapt faster to task changes in multi-agent systems. Furthermore, logical reward shaping is employed to enhance agents' cooperation and improve the interpretability of their decision-making when completing multiple tasks.

\subsection{Organization of the Paper}

The rest of this article is organized as follows. Section II introduces the preliminaries of reinforcement learning and LTL. Section III describes the algorithm model. Section IV presents the experimental design and results. Finally, the last section summarizes this work with future research directions.

\section{Preliminaries}
\label{sec2}

\subsection{Single Agent Reinforcement Learning}

The single agent reinforcement learning (SARL)  \cite{neto2005single} is based on the idea that the agent can learn to select appropriate actions by interacting with a dynamic environment and maximizing its cumulative return. This process is similar to how humans acquire knowledge and make decisions. Deep reinforcement learning has made significant progress in AI research, as demonstrated by the successes of AlphaGo \cite{silver2016mastering}, AlphaGo Zero \cite{silver2017mastering}, and AlphaStar  \cite{DBLP:journals/nature/VinyalsBCMDCCPE19}, which have shown the ability of reinforcement learning to solve complex decision-making problems.

% , and its basic architecture is shown in Figure (\ref{fig1:SARL}).

% \begin{figure}[ht]
% \centering
% \includegraphics[height=6cm,width=8cm]{image/单智能体.pdf}
% \caption{Single agent reinforcement learning process.}
% \label{fig1:SARL}
% \end{figure}

The interaction between the agent and the environment in SARL follows the Markov Decision Process (MDP). MDP is generally represented by a tuple of $\langle S, A, R, T,\gamma \rangle$. In this tuple, $S$  represents the state space of the environment. At the beginning of an episode, the environment is set to an initial state $s_0$. At timestep $t$, the agent observes the current state $s_t$. The action space is represented by $A$, and $a_t \in A$ represents the action the agent performs at timestep $t$. The function $R: S\times A\times S\rightarrow \mathbb{R}$ defines the instantaneous return of an agent from state $s_t$ to state $s_{t+1}$ through action $a_t$. The total return from the beginning time $t$ to the end of the interaction at time $K$ can be expressed as $R_t = \sum_{t'=t}^K\gamma^{t'-t}r_{t'}$. Here, $\gamma \in [0,1]$ is the discount coefficient, which is used to ensure that the later return has a more negligible impact on the reward function. It depicts the uncertainty of future returns and also limits the return function. The function $T:S\times A\times S \rightarrow [0,1]$, defines the probability distribution of the transition from $s_t$ to $s_{t+1}$ for a given action $a_t \in A$. According to the feedback reward of the environment, the agent uses positive and negative feedback to indicate whether the action is beneficial to the learning goal. The agent constantly optimizes action selection strategies through trial and error and feedback. Eventually, the agent learns a goal-oriented strategy.

\subsection{Multi-Agent Reinforcement Learning}

\begin{figure}[ht]
\centering
\includegraphics[width=8cm]{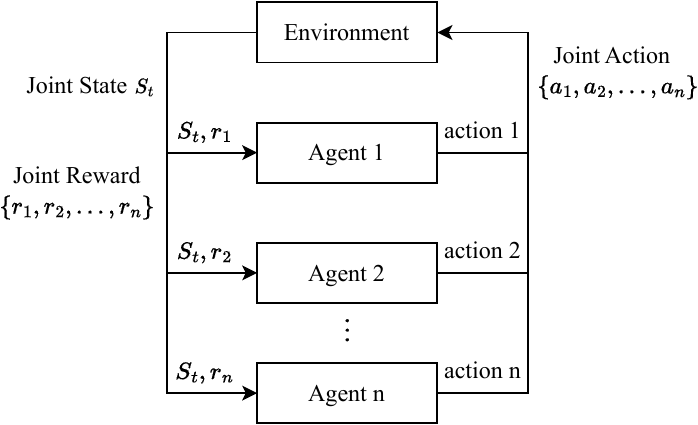}
\caption{Multi-agent reinforcement learning process.}
\label{fig2:MARL}
\end{figure}

When faced with large-scale and complex decision-making problems, a single-agent system cannot comprehend the relationship of cooperation or competition among multiple decision-makers. Therefore, the DRL model is extended to a multi-agent system with cooperation, communication, and competition among multiple agents. This is called multi-agent reinforcement learning (MARL) \cite{bucsoniu2010multi, 10004710}. The MARL framework is modeled as a Markov Game (MG):  $\langle N, S, A, R, T, \gamma \rangle$. $N$ stands for the number of agents.  $A = a_1 \times, \ldots, \times a_{N}$ is the joint action space for all agents. For $i \in [1,\ldots, N], R_i:S \times A \times S \rightarrow \mathbb{R}$ is the reward function for each agent. The reward function is assumed to be bounded. $T:S \times A \times S \rightarrow [0,1]$ is the state transition function. $\gamma$  is the discount coefficient. The multi-agent reinforcement learning scheme is shown in Figure \ref{fig2:MARL}.

\subsection{Linear Temporal Logic}

% 检查LTL的定义是否正确
Linear Temporal Logic (LTL) is a type of propositional logic with temporal modalities. Let $\mathcal{AP}$ be a set of propositions. An LTL formula consists of finite propositions $p \in \mathcal{AP}$, logical connectives $\vee$ (disjunctive), $\wedge$ (conjunctive), $\rightarrow$ (implication), $\lnot$ (negation), unary sequence operators $\bigcirc$ (next), $\square$ (always), $\diamond$ (eventually), and binary operators $\mathcal{U}$ (until), $\mathcal{R}$ (release). The following formula gives the syntax of LTL formula $\varphi$ on proposition set $\mathcal{AP}$, where $p\in \mathcal{AP}$:

\vspace{-2ex}
\begin{footnotesize}
\begin{equation}
\varphi ::= p \ |\ \lnot \varphi\ |\ \varphi \wedge \phi\ |\ \varphi \vee \phi\ | \ \bigcirc \varphi \ |\ \square \varphi \ |\ \diamond \varphi\ |\ \varphi\, \mathcal{U}\, \phi \ 
 |\ \varphi\, \mathcal{R}\, \phi. 
\label{equ1}
\end{equation}
\end{footnotesize}

These temporal operators have the following meanings: $\bigcirc \varphi$ indicates that $\varphi$ is true in the next time step; $\square \varphi$ indicates that $\varphi$ is always true; $\diamond \varphi$ indicates that $\varphi$ will eventually be true; $\varphi\, \mathcal{U} \, \phi$ means that $\varphi$ must remain true until $\phi$ becomes true; therefore, $\diamond \varphi \equiv true \, \mathcal{U}\, \varphi$. 
$\varphi\, \mathcal{R}\, \phi$ means that $\phi$ is always true or $\phi$ remains true until both $\varphi$ and $\phi$ are true.

In general, LTL describes propositions with infinite length, but this work focuses on the tasks that are completed in a finite time. Therefore, we use the co-safe Linear Temporal Logic (Co-safe LTL) \cite{10215053}, which extends LTL to specify and characterize system behaviors. Co-safe LTL specifications are typically of finite lengths, so the truth value for a given specification can be determined within a finite number of steps. For instance, the formula $\diamond \varphi_1$ meaning ``Eventually $\varphi_1$ is true" is co-safe because once $\varphi_1$ is true, what happens afterward is irrelevant. Note that $\lnot \diamond \varphi_2$ meaning ``$\varphi_2$ must always be false" is not co-safe because it can only be satisfied if $\varphi_2$ is never true in an infinite number of steps. Nonetheless, we can define a co-safe task $\lnot \varphi_2\, \mathcal{U}\, \varphi_1$, which means ``$\varphi_1$ must always be false before $\varphi_2$ is true". Thus, co-safe LTL formulae can still define some subtasks to be completed while ensuring finite attributes. In the following, unless specified, LTL formulae generally refer to co-safe LTL formulae.

Deterministic Finite Automaton (DFA) is a finite state machine that accepts or rejects a finite number of symbol strings \cite{rabin1959finite}. According to Laccrda's work \cite{DBLP:conf/ijcai/Lacerda0H15}, any co-safe formula $\varphi$ can be transformed into a corresponding DFA $\mathcal{D}_{\varphi}=\left\langle Q, \bar{q}, Q_{F}, 2^{\mathcal{AP}}, \delta_{\mathcal{D}_{\varphi}}\right\rangle$. Here, $Q$ represents a finite set of states, $\bar{q}\in Q$ is the initial state, $Q_F \subseteq Q$ is the set of final acceptance states, $2^{\mathcal{AP}}$ is the alphabet, and $\delta_{\mathcal{D}_{\varphi}}:Q\times 2^{\mathcal{AP}} \rightarrow Q$ is a transition function. The DFA $\mathcal{D}_{\varphi}$ accepts only the finite traces that satisfy $\varphi$.  The transformation process demonstrates double-exponential complexity, but it significantly improves the feasibility and automation of LTL formulae verification. Practical tools for such transformations have been developed and demonstrated to be effective in practice.

\section{Algorithm Model}

\begin{figure}[ht]
 % \begin{figure*}[htbp]
\centering
\includegraphics[width=\linewidth]{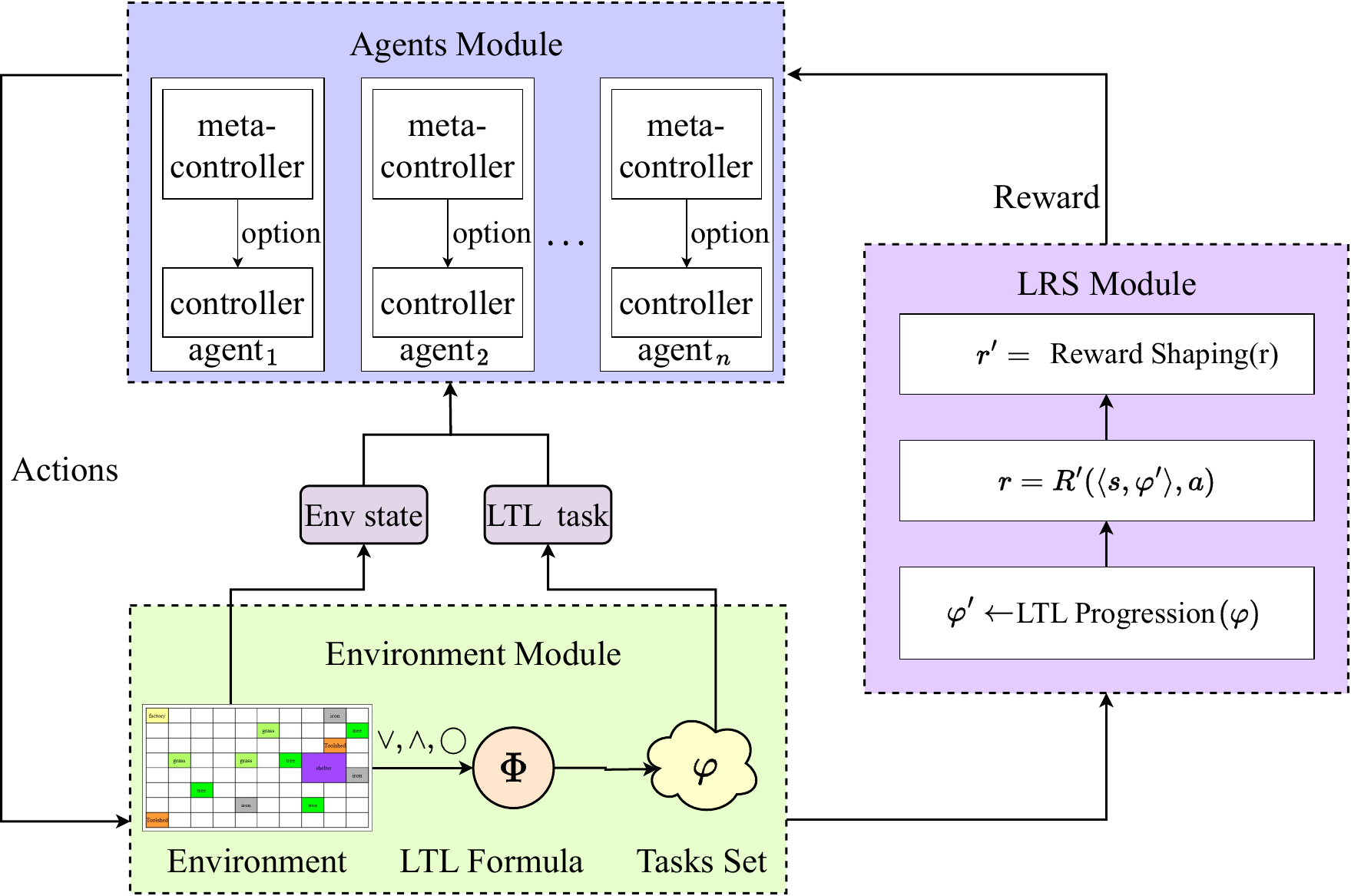}
\caption{Overall framework of MHLRS.}
\label{fig: algorithm}
% \end{figure*}
\end{figure}

The algorithm MHLRS comprises of three modules, namely, the environment module, agents module, and logic reward shaping (LRS) module. The architecture is shown in Figure \ref{fig: algorithm}. The environment consists of various components that are combined using logical and temporal connectives to create LTL formulae. These formulae act as tasks for the agents and are stored in a set called $\Phi$. Each task is represented by an LTL formula $\varphi$ that needs to be completed by the agents. Each agent in the environment adopts a two-layer hierarchical structure. The high-level meta-controller proposes strategies, which guide the agent to perform actions, while the low-level controller is responsible for executing actions to complete the target tasks. The reward shaping component assigns appropriate rewards to agents as they complete LTL tasks, after LTL progression and value iteration. To efficiently collaborate among agents, each agent can share LTL task information for the strategy learning of common tasks. It is important to note that this structure is scalable and flexible in terms of the number of agents. 

\label{sec:Alg}
% \vspace{-2ex}

The following is a detailed description of the main modules of the algorithm and the training method of MHLRS. \par

\subsection{Logic Reward Shaping Module}

Each task can be formally specified in LTL syntax. For example, Equation (\ref{equ3}) represents the task of {\it making an axe}, which is composed of four propositions and two conjunctions.

\vspace{-0.5ex}
\begin{equation}
\begin{split}
\varphi_{axe} \triangleq \diamond (got\_wood \wedge \diamond used\_workbench)\\
\wedge \diamond(got\_iron \wedge \diamond used\_factory).
\label{equ3}
\end{split}
\end{equation}

To determine the truth of the LTL formulae, a label function $L:S \rightarrow 2^\mathcal{AP}$ is introduced, where $S$ is the state set, and $\mathcal{AP}$ is the atomic proposition set. The label function $L$ can be regarded as an event detector. When some events are triggered in a state $s_t \in S$, the corresponding propositions for these events will be assigned true. We denote the set of true propositions at $s_t$ as $\sigma_t$, which is shown in Equation (\ref{equ: Label}).  

\begin{equation}
\sigma_t = L(s_t).
\label{equ: Label}
\end{equation}

For instance, in the task of  {\it making an axe} expressed by Equation (\ref{equ3}), suppose that the agent has successfully completed the subtask of {\it getting wood} at state $s_t$, then $\sigma_t= L(s_t) = \{got\_wood\}$. On the other hand, if no proposition is true in state $s_t$, $\sigma_t= L(s_t) = \emptyset$.

The truth value of the LTL formula can be determined by a sequence $\sigma$ defined as $\langle\sigma_0, \sigma_1, \sigma_2, \cdots, \sigma_t\rangle$. For any formula $\varphi$, $\langle\sigma, i\rangle\  \models \varphi$ if and only if the sequence $\sigma$ satisfies the proposition or formula $\varphi$ at time $k$, formally defined as:\par

\begin{itemize}

\item[$\bullet$] $\langle\sigma, i\rangle\  \models p$ iff $ p \in \sigma_i$, where $p \in \mathcal{AP}$.

\item[$\bullet$] $\langle\sigma, i\rangle\  \models \lnot \varphi $ iff $\langle\sigma, i\rangle\ \not \models \varphi$.

\item[$\bullet$] $\langle\sigma, i\rangle\  \models (\varphi_1 \wedge \varphi_2)$ iff $\langle\sigma, i\rangle\ \models \varphi_1$ and $\langle\sigma, i\rangle\ \models \varphi_2$.

\item[$\bullet$] $\langle \sigma, i\rangle\  \models (\varphi_1 \vee  \varphi_2)$ iff $\langle \sigma, i\rangle \ \models \varphi_1$ or $\langle \sigma, i\rangle \ \models \varphi_2$.

\item[$\bullet$] $\langle\sigma, i\rangle\  \models \bigcirc \varphi$ iff $i < t$, and $\langle\sigma, i+1\rangle\ \models \varphi$.

\item[$\bullet$] $\langle\sigma, i\rangle\  \models \square \varphi$ iff $\exists j \in [0,t]$, such that $\forall k > j,\    \langle\sigma, k\rangle\ \models \varphi$.

% \item[$\bullet$] $\langle\sigma, i\rangle\  \models \square \varphi$ iff $\forall j \in [0,t], \langle\sigma, j\rangle\ \models \varphi$.

\item[$\bullet$] $\langle\sigma, i\rangle\  \models \diamond \varphi$ iff $\exists j \in [0,t]$, $\langle\sigma, j\rangle\ \models \varphi$.

% \item[$\bullet$] $\langle\sigma, i\rangle\  \models \varphi_1\, \mathcal{U}\, \varphi_2$ iff $\exists j \in [i,t]$ such that $\langle\sigma, j\rangle\  \models \varphi_2$, and for all $k$ with $i\leq k < j$,$~\langle\sigma, k\rangle\ \models \varphi_1$ .

\item[$\bullet$] $\langle\sigma, i\rangle\  \models \varphi_1\, \mathcal{U}\, \varphi_2$ iff $\exists j \in [i,t]$ such that $\langle\sigma, j\rangle\  \models \varphi_2$, and for $\forall k$  with $k\in [i,j)$, $~\langle\sigma, k\rangle\ \models \varphi_1$.

%\item[$\bullet$] $\langle\sigma, i\rangle\  \models \varphi_1\, \mathcal{R}\, \varphi_2$ iff $\forall j \in [i,t]$ such that $\langle\sigma, j\rangle\  \models \varphi_1$ and  $\exists k$  with $i\leq k < j$, $\langle\sigma, k\rangle\ \not \models \varphi_2$. 

\item[$\bullet$] $\langle\sigma, i\rangle\  \models \varphi_1\, \mathcal{R}\, \varphi_2$ iff $\exists j\in [i,t]$ such that $\langle\sigma, j\rangle\  \models \varphi_1$ and for $\forall k$  with $k\in [i,j]$, $\langle\sigma, k\rangle\  \models \varphi_2$; or for $\forall k\in [i,t]$, $\langle\sigma, k\rangle\  \models \varphi_2$. 

\end{itemize}

If the formula is true at time $t$ (i.e. $\langle \sigma, t\rangle\  \models \varphi$ is satisfied), the agent will get a reward of $1$; and $-1$ else. Formally, given an LTL formula $\varphi$ based on a proposition set $\mathcal{AP}$, the label function $L:S \rightarrow 2^\mathcal{AP}$, as well as a sequence $\sigma = \langle \sigma_0, \sigma_1, \sigma_2,\cdots, \sigma_t\rangle$, the reward structure for $\varphi$ is defined as follows:

\begin{footnotesize}
\begin{equation}
R_{\varphi}\left(s_{0}, \ldots, s_{t}\right)=
\left\{\begin{array}{ll}
1 & \text { if } \langle \sigma, t \rangle \models \varphi, \\
-1 & \text { otherwise. } 
% \langle \sigma, t \rangle \models \neg \varphi \\
% 0 & \text { otherwise }
\end{array}\right.
\label{equ4}
\end{equation}
\end{footnotesize}
where $\sigma_i=L(s_i), i \in [0,t]$.\par

According to this reward structure, the agents receive a non-zero reward regardless of whether they complete the whole LTL task. The Q-value function of policy $\Pi$ can be defined over the state sequences, as shown in Equation (\ref{equ5}):

\begin{footnotesize}
\begin{equation}
\begin{aligned} 
Q^{\Pi}\left(\left\langle s_{0:t}\right\rangle, \mathcal{A}\right)  =
\mathbb{E}_{\Pi}\!\left[\sum_{t=0}^{\infty} \gamma^{t}\! R_{\varphi}\left(\!\left\langle s_{0:t}\right\rangle\!\right)\! \mid \!A_{t}\!=\!\mathcal{A}\right].
\label{equ5}
\end{aligned}
\end{equation}
\end{footnotesize}
\noindent where $\langle s_{0:t}\rangle$ is the abbreviation of  $\langle s_{0}, \ldots, s_{t} \rangle$. $A_t = \mathcal{A}$ is the joint actions taken by the agents in state $s_t$.\par

The goal of the agents is to learn an optimal policy $\Pi^{*}(a_t|s_t,\varphi)$ to maximize the $Q$ value. However, one challenge in the learning process is that the reward function depends on the sequence of states and is non-Markov. 
% Therefore, the learning of the tasks expressed by LTL formulas is a Non-Markovian Reward Decision Process (NMRDP) problem. 
In order to formally define this problem, we use the concept of the Non-Markovian Reward Game (NMRG). 

\begin{definition}
$\mathit{(NMRG)}$.
 A Non-Markovian Reward Game (NMRG) is modeled as a tuple $\mathcal{M} = \langle N, S, A, R_\varphi, T, \gamma \rangle$, where $N, S, A, T, \gamma$ are defined as in the MG, but the $R_\varphi$ is defined over state histories, $R_\varphi:\langle L(s_1,a_1), \ldots, L(s_t,a_t) \rangle \rightarrow \mathbb{R}$. 
\label{def1}
\end{definition}

The optimal policy $\Pi^{*}$ must consider the state sequences of the whole training process, which brings about the issue of long-term dependencies. To deal with this issue, we use the method of LTL progression, which works in the following way.

LTL progression \cite{DBLP:conf/atal/IcarteKVM18, DBLP:conf/icml/VaezipoorLIM21} is a rewriting process that retains LTL's semantics. It receives an LTL formula and the current state as input and returns a formula that needs to be further dealt with. The LTL formula is progressed according to the obtained truth assignment sequence $\{\sigma_0, \ldots, \sigma_t \}$. In the training process, the formula is updated in each step to reflect the satisfied parts in the current state. The progressed formula will only include expressions on tasks that are uncompleted. For example, the formula $\diamond(\varphi_{1} \wedge \bigcirc \diamond \varphi_2)\ $can be progressed to $\bigcirc \diamond \varphi_2$ when $\varphi_1$ is true. 

Given an LTL formula $\varphi$ and a truth assignment $\sigma_i$, the LTL progression $prog(\sigma_i,\varphi)$ on the atomic proposition set $\mathcal{AP}$ is formally defined as follows:\par

\begin{itemize}

\item[$\bullet$] $\ prog(\sigma_i,p) = true$ if $ p \in \sigma_i$, where $ p \in \mathcal{AP} $.

\item[$\bullet$]$\ prog(\sigma_i,p) = false$ if $ p \notin \sigma_i$, where $ p \in \mathcal{AP} $.

\item[$\bullet$]$\ prog(\sigma_i,\lnot \varphi) = \lnot prog(\sigma_i, \varphi)$.

\item[$\bullet$]$\ prog(\sigma_i,\varphi_1 \wedge \varphi_2) = prog(\sigma_i,\varphi_1) \wedge prog(\sigma_i, \varphi_2)$.

\item[$\bullet$]$\ prog(\sigma_i,\varphi_1 \vee \varphi_2) = prog(\sigma_i,\varphi_1) \vee prog(\sigma_i, \varphi_2)$.

\item[$\bullet$]$\ prog(\sigma_i,\bigcirc \varphi) =  \varphi$.

\item[$\bullet$]$\ prog(\sigma_i,\square \varphi) =  true$.

% \item[$\bullet$]$\ prog(\sigma_i,\diamond \varphi) =  \varphi$.

\item[$\bullet$]$\ prog(\sigma_i, \varphi_1 \mathcal{U} \varphi_2) =  prog(\sigma_i, \varphi_2) \vee (prog(\sigma_i,\varphi_1) \wedge \varphi_1 \mathcal{U} \varphi_2)$.

\item[$\bullet$]$\ prog(\sigma_i, \varphi_1 \mathcal{R} \varphi_2) =  prog(\sigma_i, \varphi_1) \wedge (prog(\sigma_i,\varphi_2) \vee \varphi_2 \mathcal{R} \varphi_1)$.

\end{itemize}

LTL progression can endow the reward function $R^{\prime}_{\varphi}$ with Markov property, shown as Equation (\ref{equ6}), and this is achieved mainly through two aspects:
(1) process the task formula $\varphi$ by progression after each action of the agents;
(2) when $\varphi$ becomes true by progression, the agent obtains the reward of $1$; $-1$ otherwise. 

% \begin{footnotesize}

\begin{equation}
R^{\prime}_{\varphi}(\langle s, \varphi\rangle, a, \langle s^{\prime}, \varphi^{\prime}\rangle)=\left\{\begin{array}{ll}
1 & \text { if } \operatorname{prog}(L(s), \varphi)=\text { true, } \\
-1 & \text { otherwise. } 
% \operatorname{prog}(L(s), \varphi)=\text { false } 
% \\
% 0 & \text { otherwise }
\end{array}\right.
\label{equ6}
\end{equation}

% \end{footnotesize}
where $s^{\prime}$ is the transition state and $\varphi^{\prime} = prog(L(s),\varphi)$.\par

By applying the $R^{\prime}_{\varphi}$ with Markov property to NMRG, the NMRG can be transformed into an MG problem that the multi-agent system can solve. For this, we apply LTL progression to the Markov Game (MG) to obtain an enhanced MG for learning LTL tasks, which is named the Enhanced Markov Game with LTL Progression (EMGLP).

\begin{definition}
$\mathit{(EMGLP)}$.
 An enhanced MG with LTL progression (EMGLP) is modeled as a tuple $\mathcal{G} = \langle N, S, A, R^{\prime}_{\varphi}, T,\\
 \mathcal{AP}, L,\Phi,\gamma \rangle$, where $N, S, A, T$, and $\gamma$ are defined as in MG or NMRG, $R^{\prime}_{\varphi}$ is the reward function to output suitable reward to the agents after acting, shown as Equation ($\ref{equ6}$), $\mathcal{AP}$ is the set of propositional symbols, $L: S \rightarrow 2^{\mathcal{AP}}$ is the label function, and $\Phi= \{\varphi_{1},\varphi_{2},\ldots,\varphi_{m}\}$, the set of tasks, is a finite non-empty set of LTL formulae over $\mathcal{AP}$.
\label{def4}
\end{definition}

Equation (\ref{equ6}) is derived from Equation (\ref{equ4}) by applying LTL progression such that $R^{\prime}_{\varphi}$ exhibits the Markov property. This allows for the learning of LTL formulae to be transformed into a reinforcement learning (RL) problem that can be handled by agents. By using $R^{\prime}_{\varphi}$ with the Markov property, the NMRG can be transformed into the EMGLP, which is a Markovian model.

\begin{definition}
$(\mathit{Transformation})$. A NMRG $\mathcal{M} = \langle N, S, A,\\
R_{\varphi}, T, \gamma \rangle$ can be transformed into an EMGLP $\mathcal{G} = \langle N, S, A,\\
R^{\prime}_{\varphi}, T, L, \Phi, \gamma \rangle$, if apply LTL progression to $R_{\varphi}$.
\label{def5}
\end{definition}

If an optimal joint strategy $\overline{\Pi}$ exists for $\mathcal{M}$, then there should also be a joint strategy $\Pi^{\prime}$ for $\mathcal{G}$ that ensures the same reward. Thus, we can find optimal strategies for $\mathcal{M}$ by solving $\mathcal{G}$ instead.

\begin{proof}
 Consider any strategy $\overline{\Pi}(\theta)$, the trajectory $\theta=\langle s_{0},\mathcal{A}_{1},s_{1},\ldots,s_{t-1},\mathcal{A}_{t},s_{t}\rangle$ for $\mathcal{M}$, where $\mathcal{A}$ is the joint action of agents. By definition of Q-value:

\begin{equation}
Q_{\mathcal{M}}^{\overline{\Pi}}\left(\langle s_{0: t}\rangle, \mathcal{A}\right)=\mathbb{E}_{\overline{\Pi}, T_{\mathcal{M}}}\left[\sum_{t=0}^{\infty} \gamma^{t} R_{\varphi}\left(\sigma_{0: t}\right) \mid A_{t}=\mathcal{A}\right].
\label{equ16}
\end{equation}
% \end{small}
where $\sigma_{i:j} = \langle \sigma_{i},\ldots,\sigma_{j} \rangle = \langle L(s_{i}, \mathcal{A}_{i}),\ldots, L(s_{j},\mathcal{A}_{j}) \rangle$. By using LTL progression, we can progress $\varphi$ to $\phi$ over the sequence $\sigma_{0:t}$, $\phi$ = $prog(\sigma_{0:t},\varphi)$, shown as Equation (\ref{equ17}):

\begin{equation}
Q_{\mathcal{M}}^{\overline{\Pi}}\left(\langle s_{0: t} \rangle, \mathcal{A}\right)=\mathbb{E}_{\overline{\Pi}, T_{\mathcal{M}}}\left[\sum_{t=0}^{\infty} \gamma^{t} R^{\prime}\left(\langle s, \phi \rangle \right) \mid A_{t}=\mathcal{A}\right].
\label{equ17}
\end{equation}

The expectation value is over $\overline{\Pi}$ and $T_{\mathcal{M}}$ that is defined in terms of $\mathcal{M}$. However, we can construct an equivalent strategy $\Pi^{\prime}$ for $\mathcal{G}$, where $\overline{\Pi}(\theta) = \Pi^{\prime}(\mathcal{A}^{\prime}_{1},\ldots,\mathcal{A}^{\prime}_{t},s^{\prime}_{t})$. What's more, $T_{\mathcal{M}}(s_{t},\mathcal{A},s_{t+1})$ is equivalent to $T_{\mathcal{G}}(s^{\prime}_{t},\mathcal{A}^{\prime},s^{\prime}_{t+1})$ for every $\mathcal{A}^{\prime} \in \mathbb{A}$ ($\mathbb{A}$ is the joint action set) . Replace $\overline{\Pi}$ by $\Pi^{\prime}$ and $T_\mathcal{M}$ by $T_{\mathcal{G}}$ in the expectation value, we will get the following equivalence: $Q_{\mathcal{M}}^{\overline{\Pi}}(s_{0:t},\mathcal{A}) = Q_{\mathcal{G}}^{\Pi^{\prime}}(s_{0:t}^{\prime},\mathcal{A}^{\prime})$ for any strategy $\overline{\Pi}$ and trajectory $\langle s_{0},\mathcal{A}_{1},s_{1},\ldots,s_{t-1},\mathcal{A}_{t},s_{t}\rangle$, shown as Equation (\ref{equ18}):

% \begin{small}
\begin{equation}
Q_{\mathcal{G}}^{\Pi^{\prime}}\left(\langle s_{0: t}^{\prime} \rangle, \mathcal{A}^{\prime}\right)=\mathbb{E}_{\Pi^{\prime}, T_{\mathcal{G}}}\left[\sum_{t=0}^{\infty} \gamma^{t} R^{\prime}\left(\langle s^{\prime}, \phi \rangle \right) \mid A_{t}=\mathcal{A}^{\prime}\right].
\label{equ18}
\end{equation}
% \end{small}

In particular, if $\Pi_{*}^{\prime}$ is optimal for $\mathcal{G}$, then $Q_{\mathcal{M}}^{\overline{\Pi}_{*}}(s_{0:t},\mathcal{A}) = Q_{\mathcal{G}}^{\Pi_{*}^{\prime}}(s_{0:t}^{\prime},\mathcal{A}^{\prime}) \geq Q_{\mathcal{G}}^{\Pi^{\prime}}(s_{0:t}^{\prime},\mathcal{A}^{\prime}) = Q_{\mathcal{M}}^{\overline{\Pi}}(s_{0:t},\mathcal{A})$ for any strategy $\Pi^{\prime}$ and joint actions $\mathcal{A}^{\prime}$. Therefore, $\overline{\Pi}_{*}$ is optimal for $\mathcal{M}$.
\end{proof}

Now, we have proved that if the optimal joint strategy $\Pi^{\prime}_{*}$ of $\mathcal{G}$ is obtained, it is equivalent to obtaining the optimal strategy $\overline{\Pi}_{*}$ for $\mathcal{M}$.

In order to enhance agents' coordination, a value iteration method is used to evaluate the actions taken by each agent. If an agent acquires a portion of the raw materials at state $s$ that brings it closer to completing the task, it will receive a relatively high evaluation value.  Conversely, if an action causes the agent to deviate from the task direction, it will receive a low evaluation value.

\begin{equation}
V(s^{\prime}) = \xi(R(s,a,s^{\prime}) + \gamma V(s|A_t =a)).
\label{equ16:V}
\end{equation}

In Equation (\ref{equ16:V}), $s^{\prime}$ is the transitioned state and $V(s^{\prime})$ is the evaluation value after the agent acts an action $a$ at state $s$. The initial evaluation value $V(s_0)$ for each agent is a small constant. $\xi$ is state transition probability. $R(s,a,s^{\prime})$ is the reward function, as shown in Equation (\ref{equ6}).

After executing the actions, each agent will receive an evaluation value denoted by $V_j, j \in(1, \ldots, n)$, where $n$ denotes the number of agents in the environment. We select the highest evaluation value $V_{max}$ and the lowest evaluation value $V_{min}$ from the set $[V_1, V_2, \ldots, V_n]$. The reward function is defined based on the evaluation values $V_{max}$ and $V_{min}$ using the concept of reward shaping:

\begin{equation}
R^{\prime}(s,a,s^{\prime}) = R(s,a,s^{\prime}) + V_{min} - \gamma V_{max}.
\label{equ12}
\end{equation}

The intuition of reward shaping is to find an appropriate potential function to enhance the structure of the reward function and make it easier to learn policies. In this work, instead of using a potential function, the evaluation value after the agent's action is used. Note that $\gamma$ in Equations (\ref{equ16:V}) and (\ref{equ12}) is a hyperparameter close to 1, typically 0.9. Therefore, the value of $V_{min}- \gamma V_{max}$ is negative in most cases. Adding this negative feedback will enable agents to learn how to cooperate with each other to accomplish tasks. When $V_{min}- \gamma V_{max} = 0$, it means that the agents are in an optimal state of collaboration. This reward structure helps the agents to collaborate and achieve their goals.

\subsection{Hierarchical Structure of Single Agent}

This section explains the hierarchical reinforcement learning architecture of each agent. The algorithm is based on the framework of options. An option is a hierarchical reinforcement learning method proposed by Sutton \cite{sutton1999between}. An option provides a policy of subgoal, guiding an agent to achieve specific tasks or sub-objectives through a sequence of actions. 

An option for learning a proposition $p$ is defined by $<I_p, \pi_p, T_p>$. Here, $I_p \subseteq S$ represents the set of initial states of the option. $\pi_p$ is the policy that defines the actions to be taken by the agent and is generally expressed as $\pi:S \times A \rightarrow [0,1]$. $T_p$ represents the set of states where the option terminates. In our algorithm, an option terminates either when $p$ is true or when the agent reaches the maximum steps.

% \vspace{-0.2cm}
\begin{figure}[ht]
\centering
\includegraphics[height=5cm,width=6cm]{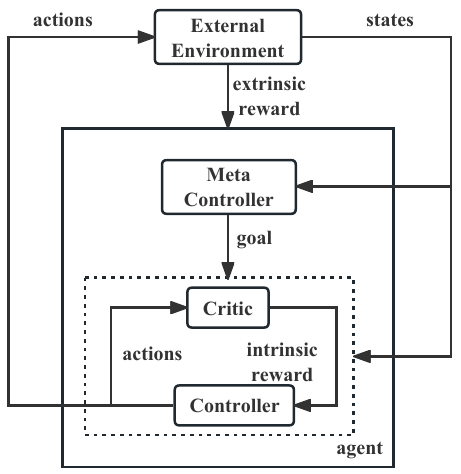}
\caption{Hierarchical structure.}
\label{fig: structure}
\end{figure}

% \vspace{-0.2cm}

Each agent adopts a two-stage hierarchical structure composed of a meta-controller and a controller, as shown in Figure \ref{fig: structure}. The meta-controller serves as the high-level component which receives state $s_t$ and generates options, defined as the policies $\pi_g$ for each subgoal $g$. Each subgoal is to satisfy a proposition in the LTL task. Once one subgoal $g_t$ is selected, it remains unchanged for a certain time step until it is realized or terminated. The agent also includes a critic that assesses whether the subgoal is achieved, and then the reward function provides appropriate rewards to the agent. Unlike the criticism in the actor-critic architecture, this critic is not a neural network and does not evaluate the controller's actions. 

The low-level component of the agent is a controller modeled by Double Deep Q-Learning (DDQN) \cite{van2016deep}. It is responsible for executing actions by following the option generated by the meta-controller and receives rewards based on the completion of tasks.  The objective of the controller is to maximize the cumulative intrinsic reward, defined as $R_{t}(g)=\sum_{t}^{\infty}  r_{t(g)}, t \ge 0$. When the controller achieves the goal, the intrinsic reward $r$ is 1; otherwise, it is 0. The meta-controller's objective is to optimize the cumulative external rewards, defined as $F_{t}=\sum_{t}^{\infty}f_t$, where $f_t$ is the reward from the environment, as shown in Equation (\ref{equ12}). \par

The agent in the hierarchical structure uses two different replay buffers - $D_1$ and $D_2$ - to store experience. $D_1$ stores the experience of the controller, which includes $(s_t,a_t,g_t,r_t,s_{t+1})$, and it collects experience once at each time step of the low-level policy. On the other hand, $D_2$ stores the experience of the meta-controller, which includes $(s_t,g_t,f_t,s_{t+Z})$, and it collects experience at each $Z$ step or when the goal is reselected. To approximate the optimal value $Q^{*}(s,g)$, DDQN with parameter $(\theta_1, \theta_2)$ is used. The parameter $(\theta_1,\theta_2)$ is updated periodically from the $(D_1, D_2)$ of the replay buffer pool without affecting each other.

\subsection{Training of MHLRS}

The training process of MHLRS is presented in Algorithm (\ref{alg1}). 

\begin{breakablealgorithm}
% \begin{algorithm}[H]
	\caption{Training process of MHLRS}%算法标题
	\begin{algorithmic}[1]%一行一个标行号
            
            \STATE Initialize replay buffers \{$D_1$, $D_2$\} and parameters \{$\theta_1$, $\theta_2$\} for each agent's controller and meta-controller.                       
		\STATE Initialize exploration probability $\epsilon_{1,g} = 1$ for each agent's controller for all goals $g$ and $\epsilon_{2} = 1$ for each agent's meta-controller.
                       
            \FOR{$j = 1, num\_episodes$}
            \STATE $\varphi \leftarrow$ CurriculumLearner($\Phi$)
            \STATE $s \leftarrow $ GetInitialState()
            % \STATE $g \leftarrow $ EPSGREEDY($s, \mathcal{G}, \epsilon_2, Q_2$)
            \IF{$random() < \epsilon$}
            \STATE $g \leftarrow$ random element from set $\mathcal{G}$
            \ELSE 
            \STATE $g \leftarrow argmax_{g_{t}\in \mathcal{G}}Q(s, g_{t})$
            \ENDIF          
            \WHILE{$\varphi \notin \{true, false\}$ and not EnvDeadEnd(s)}
            \STATE F $\leftarrow 0$
            \STATE $s_{0} \leftarrow s$
            \WHILE{not (EnvDeadEnd(s) or goal $g$ reached)}
            \WHILE{$i < n$}
            \STATE $a \leftarrow$ agent$_i$.GetActionEpsilonGreedy($Q_{\varphi}, s$)
            \STATE Execute $a$ and obtain next state $s^{\prime}$ and extrinsic reward $R_{i}(s, a, s^{\prime})$ from environment
            \STATE Obtain intrinsic reward $r_t$ from internal critic
            \STATE $V_i = \xi(R_{i}(s, a, s^{\prime}) + \gamma V(s|a))$
            \STATE Store transition ($\{s, g\}, a, r_t, \{s^{\prime}, g\}$) in $\mathcal{D}_1$
            \STATE agent$_i$.UPDATEPARAMS($\mathcal{L}_1(\theta_{1, j})$, $\mathcal{D}_1$)
            \STATE agent$_i$.UPDATEPARAMS($\mathcal{L}_2(\theta_{2, j})$, $\mathcal{D}_2$)
            \STATE $s \leftarrow s^{\prime}$
            \STATE i += 1
            \ENDWHILE
            \STATE $F_i \leftarrow R_{i}(s, a, s^{\prime}) + V_{min} - \gamma V_{max}$
            \STATE Store transition ($s_0, g, F_i, s^{\prime}$ in $\mathcal{D}_2$)                       
            \ENDWHILE
            \IF{s is not terminal}
            \STATE $\varphi \leftarrow$ prog($L(s), \varphi$)
            \IF{$random() < \epsilon$}
            \STATE $g \leftarrow$ random element from set $\mathcal{G}$
            \ELSE 
            \STATE $g \leftarrow argmax_{g_{t}\in \mathcal{G}}Q(s, g_{t})$
            \ENDIF
            \ENDIF
            \ENDWHILE
            \STATE Anneal $\epsilon_2$ and $\epsilon_1$
            \ENDFOR
	\end{algorithmic}
\label{alg1}
% \end{algorithm}
\end{breakablealgorithm}

The environment is comprised of a set of tasks called $\Phi$, which is made up of LTL formulae. Each agent initializes its meta-controller and controller. Once initialization is complete, a curriculum learner \cite{leon2020extended} selects tasks from the task set. Suppose the task set $\Phi$ contains $m$ tasks: $\varphi_0, \varphi_1, \ldots, \varphi_{m-1}$. The curriculum learner selects tasks based on the given task order and tracks the success rate of each task in training:

\begin{equation}
P_{succ}(i)\! = \! \it{Num}\_\it{Succ}(\varphi_i) / \it{Num}(episodes).
\label{equ11}
\end{equation}

\noindent where $i \in [0,m\!-\!1]$. $P_{succ}(i)$ quantifies the proficiency of the agent in task $\varphi_i$. If $P_{succ}(i) < \varepsilon$, the task $\varphi_{i}$ can be selected according to the order. If $P_{succ}(i) \geq \varepsilon$, then the next task of $\varphi_{i+1}$ can be selected. Here, $\varepsilon$ is a threshold close to 1 (e.g., 0.98). The reason is that when the success rate of a task is close to 1, it indicates that the agent has mastered the method to solve it. So, more learning opportunities will be given to tasks with a low success rate, which is more conducive to training the agent to complete all tasks.

The agents start learning and solving a selected task, beginning from an initial state $\varphi_0$. They share information about the task with each other and propose the best options to complete it based on the current environment state. After executing actions, each agent receives a reward based on Equation (\ref{equ12}). The Q-value function is updated, and the experience is stored in the replay buffer during each state transition. Once the current task is completed, the curriculum learner selects the next task using Equation (\ref{equ11}). This process is repeated until the end of the training.

\section{Experiments}

We conducted experiments on a grid map similar to Minecraft, which was suggested by \cite{andreas2017modular}. This map is well-suited for a formal language that represents environmental tasks and is widely used in the literature. In order to test the effectiveness of our proposed multi-agent learning method, we expanded the environment to include multiple agents. These agents worked together to complete multiple tasks.

\begin{figure}[ht]
\centering
\includegraphics[width=8.8cm]{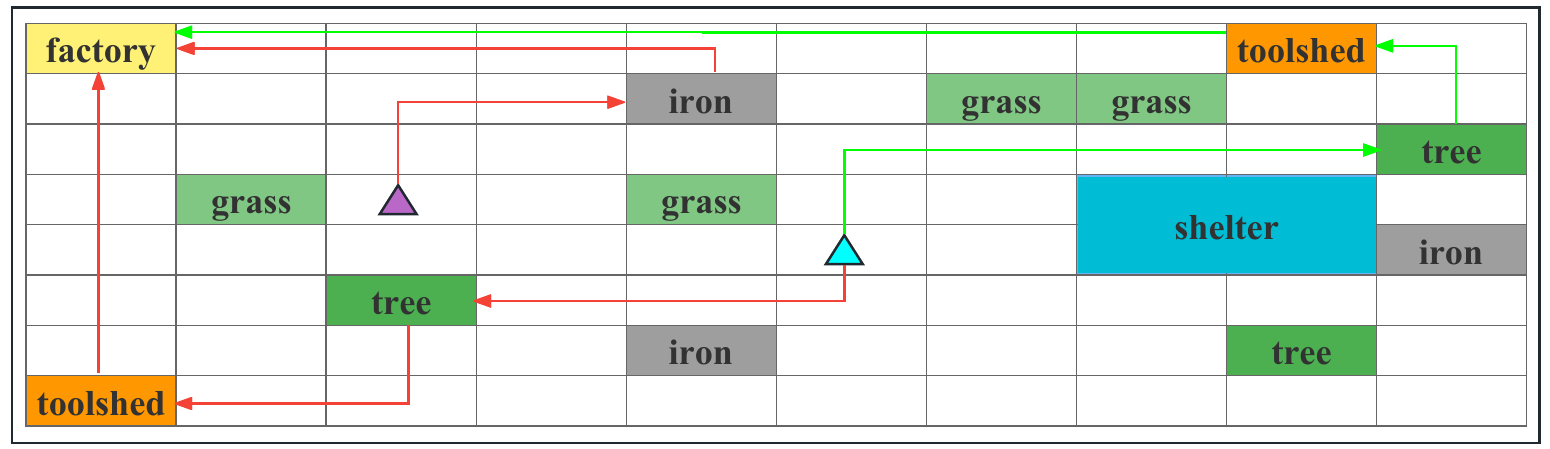}
% \includesvg[height=5cm,width=7cm]{MLP}
\caption{Minecraft-like Map.}
\label{fig:map}
\end{figure}

As shown in Figure \ref{fig:map}, there are two agents represented by purple and cyan triangles. Also, there are various raw materials and tools. There are a total of 10 tasks that need to be completed, such as making axe, fishing rod, etc.  These tasks can be expressed as $\Phi = \{\varphi_{axe},\varphi_{fishing\_rod},\ldots\}$. The strategy to complete these tasks may not be unique, and the agents have to learn the optimal one. For instance, to make an axe, one agent can look for iron,  while the other can search for wood and toolshed. They can meet at the factory and make the axe. The red trajectory in the map shows the optimal path to complete the tasks with the least number of steps. The green trajectory is also feasible but not optimal. The existence of a sub-optimal path increases the difficulty of learning.

\subsection{Baseline Algorithms}

We use three state-of-the-art algorithms as the baselines. The first baseline is the FALCON algorithm based on the commander-unit structure proposed by \cite{DBLP:journals/eswa/ZhouSTO21}. This hierarchical structure includes a high-level commander model and a multiple-agents model for low-level tasks. The low-level model accepts the command of commanders and makes its decisions. It has an excellent performance in real-time strategy games. 

The second baseline is I-LPOPL \cite{leon2020extended}, which is a multi-agent extension of LPOPL \cite{DBLP:conf/atal/IcarteKVM18} and has been specifically designed to use LTL specifications to learn multi-tasks. I-LPOPL combines LPOPL with Independent Q-learning to obtain an algorithm that can handle LTL tasks in a multi-agent environment. 

The third baseline is the I-DQN-L \cite{leon2020extended}, which extends the multi-agent reinforcement learning algorithm I-DQN \cite{tampuu2017multiagent} with LTL. Each agent is trained with an independent DQN, learns to follow the LTL specification, and uses the reward function designed by LTL instead of the classic reward function.

\subsection{Experimental Setup}
\subsubsection{Maps}

Environmental maps consist of two types: random maps and adversarial maps. Random maps have raw materials and agent locations generated randomly. Adversarial maps, on the other hand, are designed to test the effectiveness of agent collaboration. They have the highest ratio between locally and globally optimal solutions among 1000 randomly generated maps. Each algorithm runs independently three times on each map. After 1000 steps on each map, the target policy will be tested on all given tasks.

\subsubsection{Hyperparameters and Features}

All of the tested algorithms consider the same features, actions, network architecture, and optimizer. The input features are stored in a vector that records the distance from each object to the agents. The implementation of DQN and DDQN is based on the OpenAI benchmark \cite{hesse2017openai}. We use a feedforward network with ReLu as the activation function, which has two hidden layers and 64 neurons. The network uses the Adam optimizer with a learning rate of 0.0005. Sampling is done with a batch size of 32 transitions over the replay buffer of size 25,000, and the target network is updated every 100 steps. The discount factor is 0.9. For the curriculum learner, $\epsilon$ is set to 0.98.

\subsubsection{The Format of Experimental Results}

The experiments are divided into two categories. One is to compare the performance of our algorithm with the baseline algorithms, and the other is the ablation experiment to test the effect of LTL and reward shaping in the algorithm.

The experimental results of the four algorithms on two types of maps, as well as the ablation experiments, are presented in a unified format. The maximum and average rewards obtained by each algorithm in the test are shown in the tables. The maximum reward is the highest reward obtained when completing all tasks during training, while the average reward is the sum of all test rewards divided by the number of test times. In all the figures for experimental results, the purple line represents MHLRS, the green line represents FALCON, the yellow line represents I-LPOPL, and the red line represents I-DQN-L. The cyan line represents MHLRS-RS (MHLRS without reward shaping), while the black line represents MHLRS-LTL (MHLRS without LTL). The X-axis represents the number of training steps, and the Y-axis represents the average reward for all tasks during training. If a task is completed, the agent gets a reward of $+1$ and $-1$ otherwise. If the reported average reward is positive, it means that the number of completed tasks is greater than the number of failed tasks out of the total of 10 tasks.

To evaluate the performance of MHLRS, we conducted three experiments with different task settings.

\subsection{Experiment 1: Sequential Sub-Tasks}

In this experiment, there are 10 tasks in Minecraft that need to be completed. These tasks are transformed into LTL formulae, which contain a sequence of attributes to be implemented. For example, the task $``make\ $ $ fishing\_rod"$ can be defined by the propositions: $got\_$
$wood,\ used\_toolshed, \ got\_grass,\ used\_workbench$. The toolshed is used to make fishing wires by connecting grasses (Grasses can be processed into ropes as fishing wires after using the toolshed, which is just a simplified rule), and the wood is made into a rod by the workbench. This task can be transformed into the following formula:

\begin{equation}
\begin{aligned}
\varphi_{fishing\_rod}\triangleq \diamond(got\_wood \wedge \diamond(used\_workbench\\
\wedge \diamond(got\_grass \wedge \diamond used\_toolshed))).
\label{equ13}
\end{aligned}
\end{equation}
According to the formula, agents must obtain raw materials to complete tasks.

\begin{table}[htbp]
\caption{Rewards in the sequential tasks.}\label{table1}
\centering
\resizebox{\linewidth}{10mm}{
\setlength{\tabcolsep}{1mm}
\begin{tabular}{c| c| c| c| c}
\hline
\multirow{2}{*}{Algorithm} & \multicolumn{2}{c|}{random map} & \multicolumn{2}{c}{adversarial map}\\
\cline{2-3}\cline{4-5}

& maximum reward & average reward& maximum reward & average reward\\
 
\hline
MHLRS	& \textbf{8.02} &	\textbf{3.81} & \textbf{8.0} &	\textbf{3.11}\\

FALCON 	& 	1.71 &	-4.66 & 	-3.68 &	-7.59\\

I-LPOPL 	& 	-6.72 & 	-7.57  &	-6.48 & 	-6.48\\

I-DQN-L 	& 	-7.6 &	-8.07 & 	-7.83 &	-8.34 \\
\hline
\end{tabular}}
\end{table}

\iffalse

\begin{figure}[htbp]
\begin{minipage}[t]{0.5\linewidth}
    \includegraphics[height=4cm,width=\linewidth]{image/se1.jpg} 
    % \caption{caption1} # 修改描述
    % \label{f1}
\end{minipage}%
    \hfill%
\begin{minipage}[t]{0.5\linewidth}
    \includegraphics[height=4cm,width=\linewidth]{image/se2.jpg}
    % \caption{caption2}
    % \label{f2}
\end{minipage} 
\caption{Results of average rewards in Experiment 1.}
\label{fig5}
\end{figure}
\fi

\begin{figure}[htbp]
 \centering
 \includegraphics[height=4.5cm]{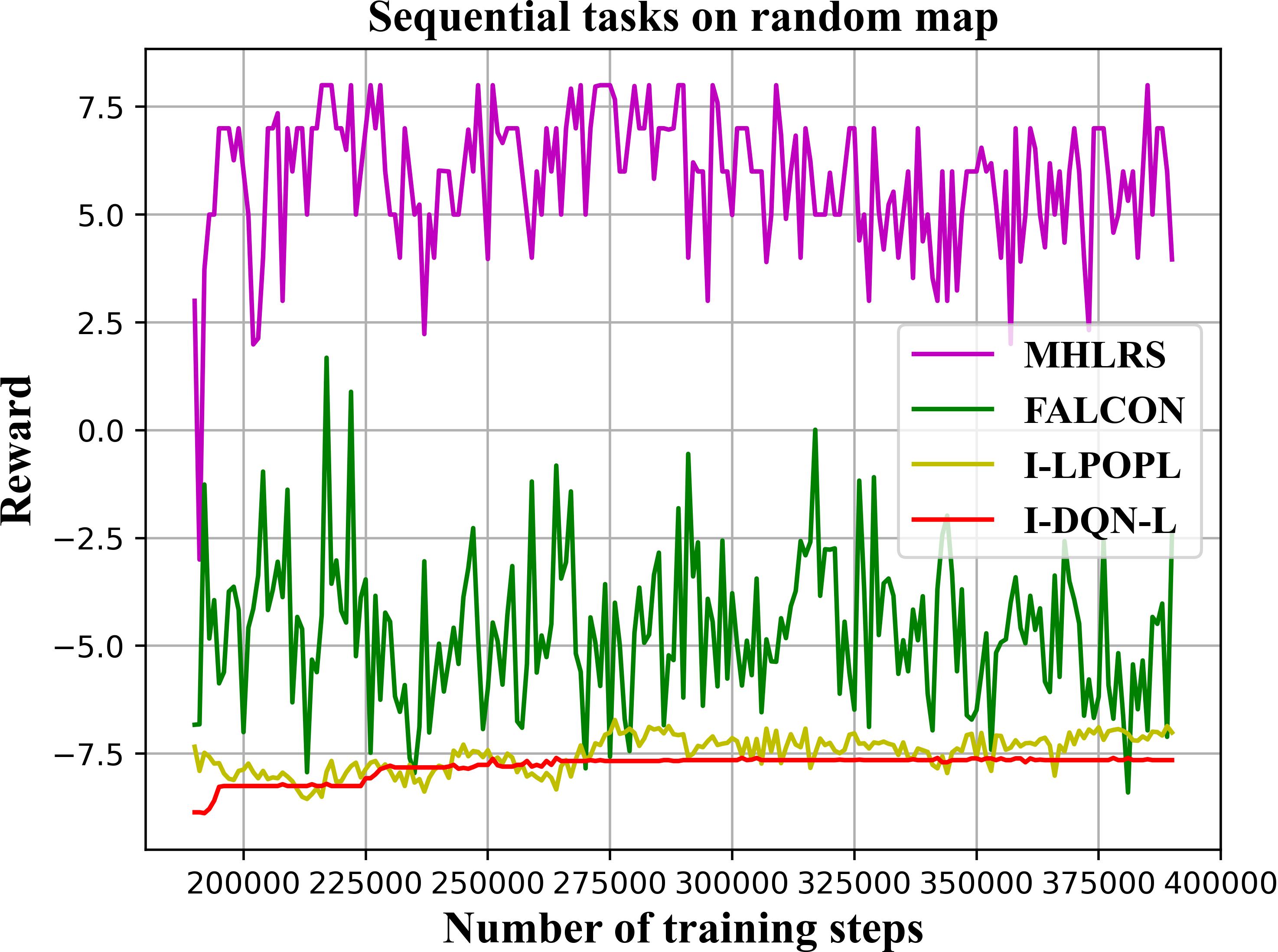}
 \caption{\vspace{-1ex}Reward curves in sequential tasks (random map).}
 \label{fig:sequence-random}
 \end{figure}

\begin{figure}[htbp]
 \centering
 \includegraphics[height=4.5cm]{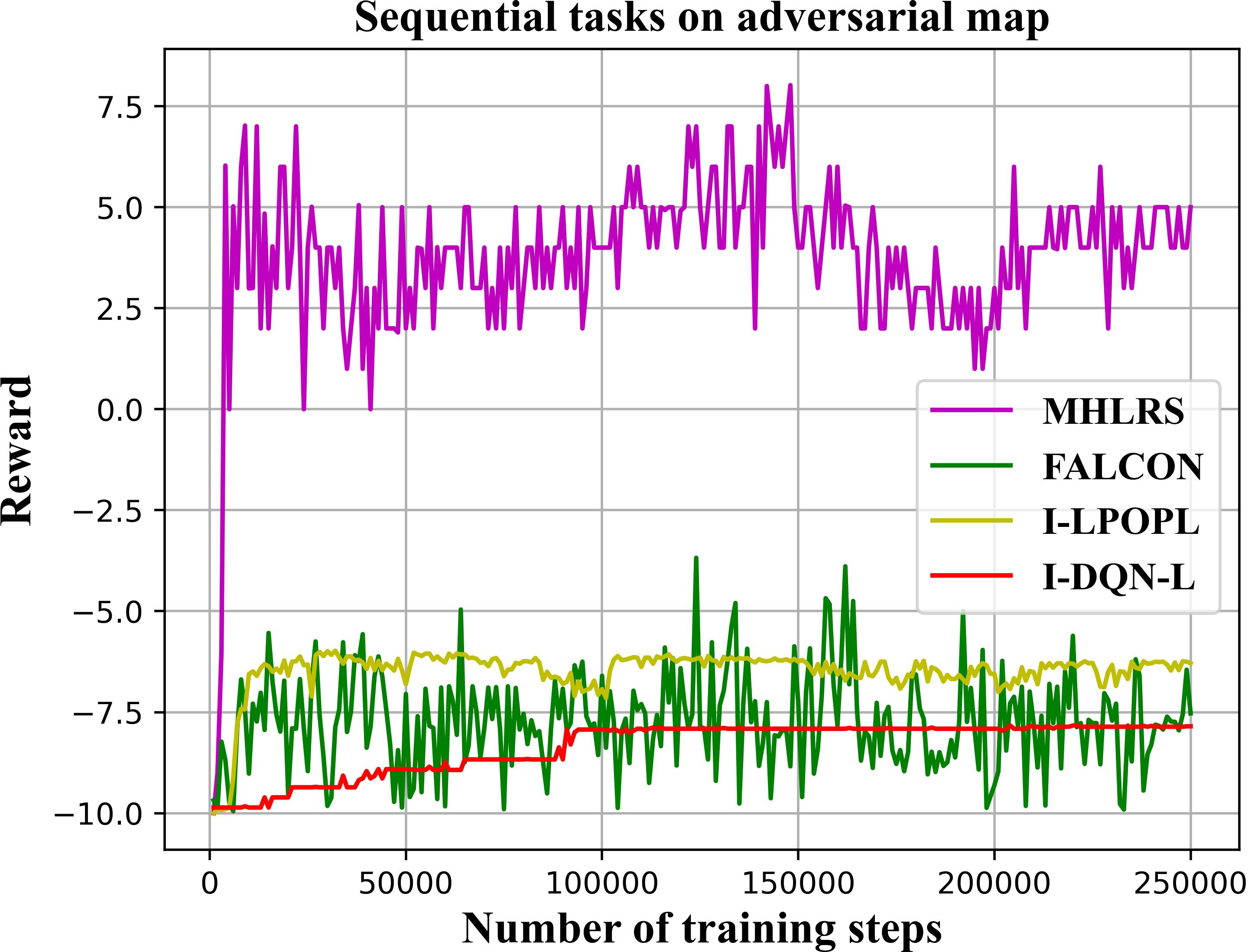}
 \caption{Reward curves in sequential tasks (adversarial map).}
 \label{fig:sequence-adversarial}
 \end{figure}

Figures \ref{fig:sequence-random}-\ref{fig:sequence-adversarial} show the curve of the average reward obtained by each algorithm when performing 10 tasks on the two types of maps. The values of the maximum reward and average reward are presented in Table \ref{table1}.  According to the results, the FALCON  algorithm obtains higher rewards than the other two baseline algorithms on random maps. However, the rewards obtained are highly fluctuating. Especially in adversarial maps, the rewards of FALCON are considerably reduced. The rewards of I-LPOPL on the two types of maps are not positive and are lower than -6. I-DQN-L obtains the lowest reward. Compared with the three baseline algorithms, MHLRS performs exceptionally well and achieves the highest reward, exceeding that of the three baseline algorithms. The maximum reward is over 8, and the average reward is over 3. Although the reward obtained on adversarial maps has an inevitable decline, it still converges to 5.0 at last, indicating that MHLRS can coordinate multi-agents to learn and gain a better policy than the baseline.

\begin{figure}[htbp]
 \centering
 \includegraphics[height=4.5cm]{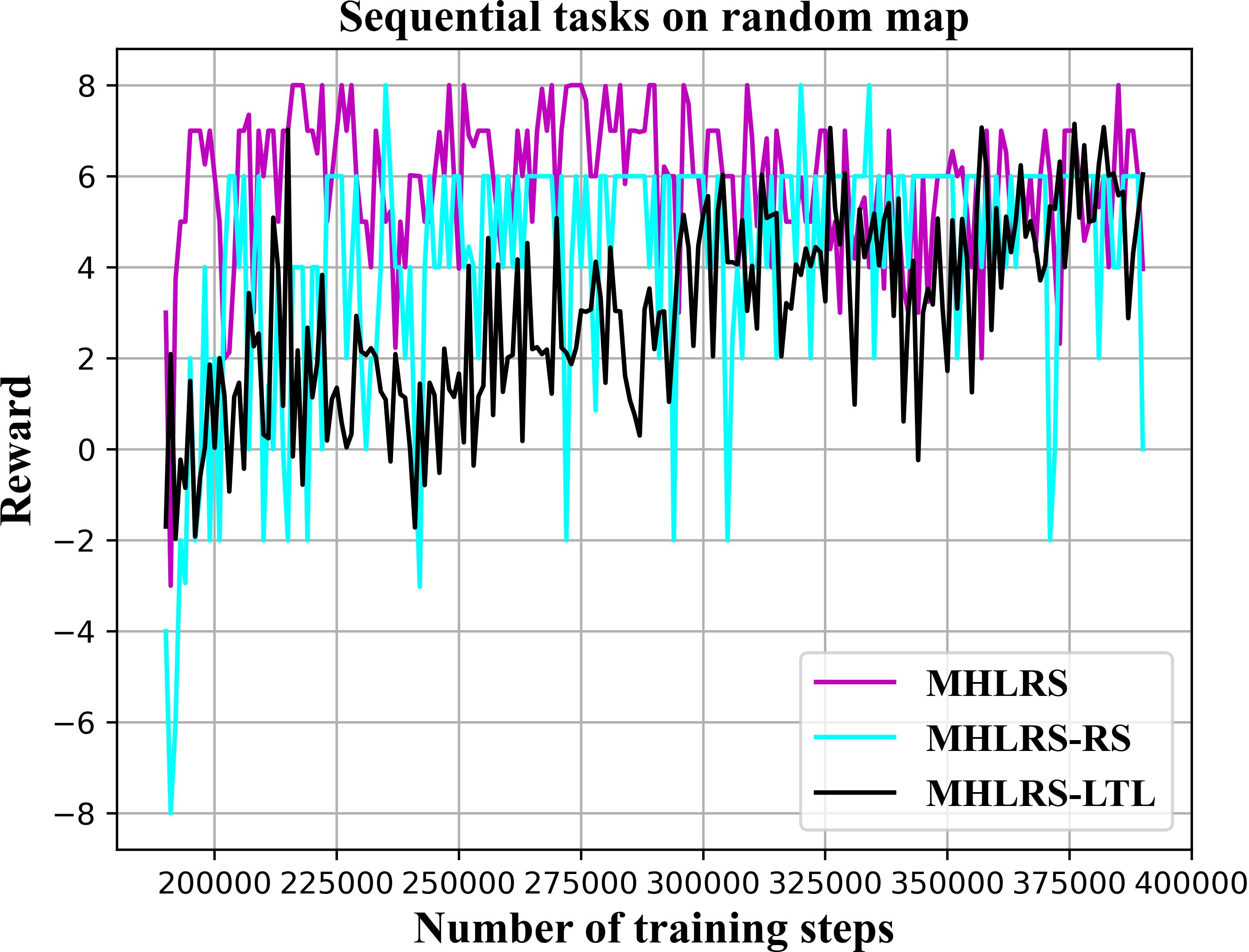}
 \caption{\vspace{-1ex}Reward curves of ablation experiment in sequential tasks (random map).}
 \label{fig:sequence-random-ablation}
 \end{figure}

\begin{figure}[htbp]
 \centering
 \includegraphics[height=4.5cm]{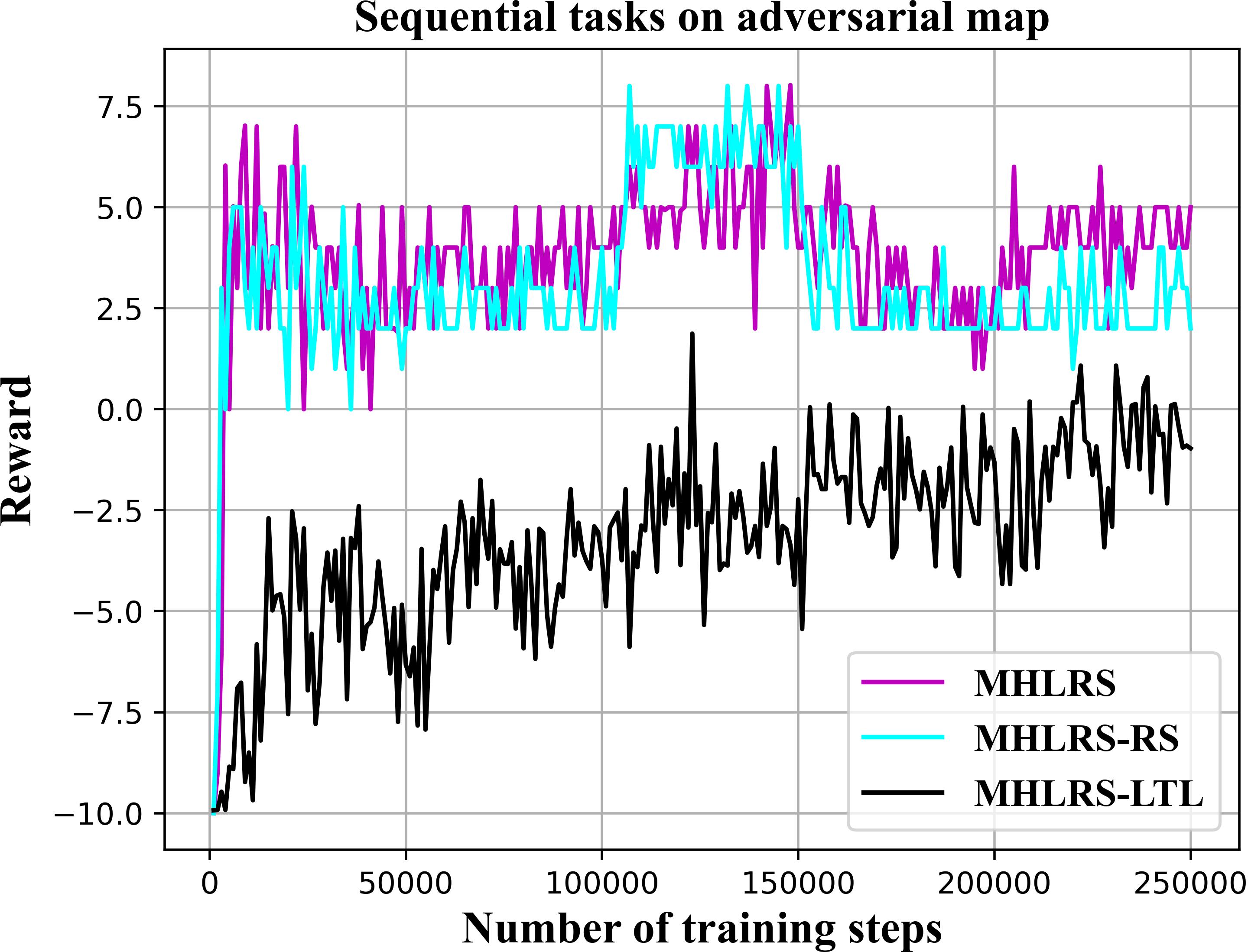}
 \caption{Reward curves of ablation experiment in sequential tasks (adversarial map).}
 \label{fig:sequence-adversarial-ablation}
 \end{figure}

 To show the effects of the design of the reward shaping and LTL design, separate experiments were conducted, and the results are presented in  Figures \ref{fig:sequence-random-ablation}-\ref{fig:sequence-adversarial-ablation} and Table \ref{table1-ablation}. According to the ablation experiment, in both two types of maps, the MHLRS-RS has a maximum reward that is close to the MHLRS. However, the algorithm becomes more unstable, and the average reward is lower than the MHLRS. This illustrates the effectiveness of reward shaping in promoting the learning cooperation strategy among agents and improving the average reward during task completion. On the other hand, the MHLRS-LTL shows significant performance degradation. The average rewards and maximum rewards are much lower than MHLRS on both random and adversarial maps, highlighting the importance of LTL in the algorithm.

\begin{table}[htbp]
\caption{Rewards of ablation experiment in the sequential tasks.}\label{table1-ablation}
\centering
\resizebox{\linewidth}{8.5mm}{
\setlength{\tabcolsep}{1mm}
\begin{tabular}{c| c| c| c| c}

\hline
\multirow{2}{*}{Algorithm} & \multicolumn{2}{c|}{random map} & \multicolumn{2}{c}{adversarial map}\\
\cline{2-3}\cline{4-5}

& maximum reward & average reward& maximum reward & average reward\\
 
\hline
MHLRS	& \textbf{8.02} &	\textbf{3.81} & \textbf{8.01} &	\textbf{3.11}\\

MHLRS-RS 	& 	7.98 &	3.2 & 	8.0 &	2.37\\

MHLRS-LTL 	& 	7.15 & 	0.91  &	1.87 & 	-3.19\\

\hline
\end{tabular}}
\end{table}

\subsection{Experiment 2: Interleaving of Sub-Tasks}
In Experiment 1, the order of completing sub-tasks is predetermined. The agents follow the LTL formulae to accomplish these tasks, which provides guidance for the agents and thus reduces the difficulty. Whereas, in general cases,  subtasks can be completed in different orders. For example, in the task $\varphi_{fishing\_rod}$,  the agent needs to create a fishing wire made of grass and a rod made of wood.  The order of making these two materials can be arbitrary. Therefore,  we have rephrased the sequence-based tasks and removed expressions that specify the order of the formulae that are unnecessary.  For example, $\varphi_{fishing\_rod}$ is rewritten as:

\begin{small}
\begin{equation}
\begin{aligned}
\diamond(got\_wood  \wedge \diamond(used\_toolshed  \wedge \diamond  used\_workbench))\\
{\wedge \diamond  (got\_grass  \wedge \diamond  used\_workbench).}
\end{aligned}
\label{equ14}
\end{equation}
 \end{small}

 \begin{table}[htbp]
\caption{Rewards in the interleaving tasks.}\label{table3}
\centering
\resizebox{\linewidth}{10mm}{
\setlength{\tabcolsep}{1mm}
\begin{tabular}{c| c| c| c| c}
\hline
\multirow{2}{*}{Algorithm} & \multicolumn{2}{c|}{random map} & \multicolumn{2}{c}{adversarial map}\\
\cline{2-3}\cline{4-5}

& maximum reward & average reward& maximum reward & average reward\\
 
\hline
MHLRS	& \textbf{10.0} &	\textbf{4.71} & \textbf{9.0} &	\textbf{4.11}\\

FALCON 	& 	0.02 &	-4.54	& 	-3.61 &	-7.00\\

I-LPOPL 	& 	-5.54 & 	-6.52 & 	-5.27 & 	-5.97\\

I-DQN-L 	& 	-6.48 &	-7.48	& 	-7.14 &	-8.12\\
\hline
\end{tabular}}

\end{table}
 
\begin{figure}[htbp]
 \centering
 \includegraphics[height=4.5cm]{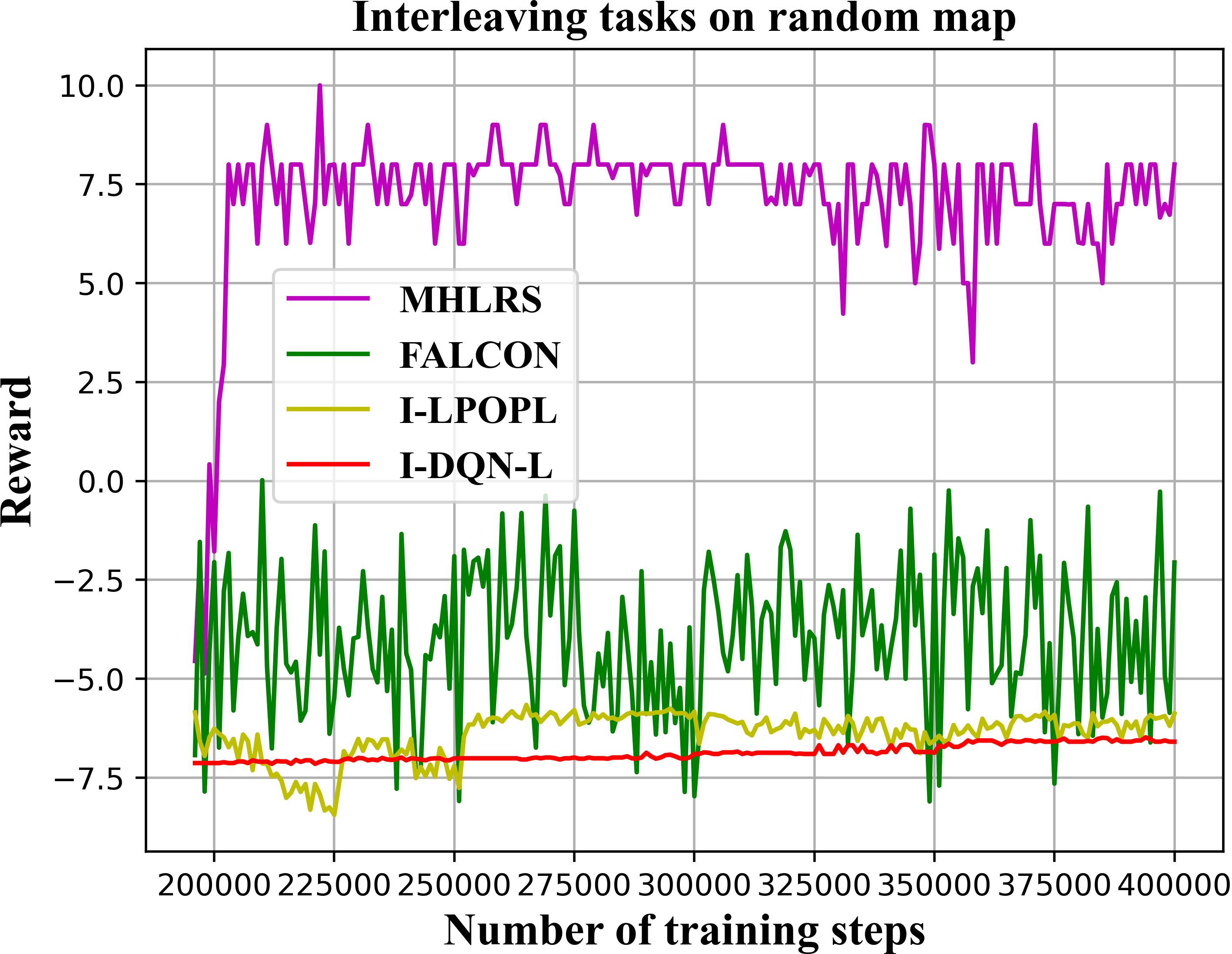}
 \caption{Reward curves in interleaving tasks (random map).}
 \label{fig:Interleaving-random}
 \end{figure}

\begin{figure}[htbp]
 \centering
 \includegraphics[height=4.5cm]{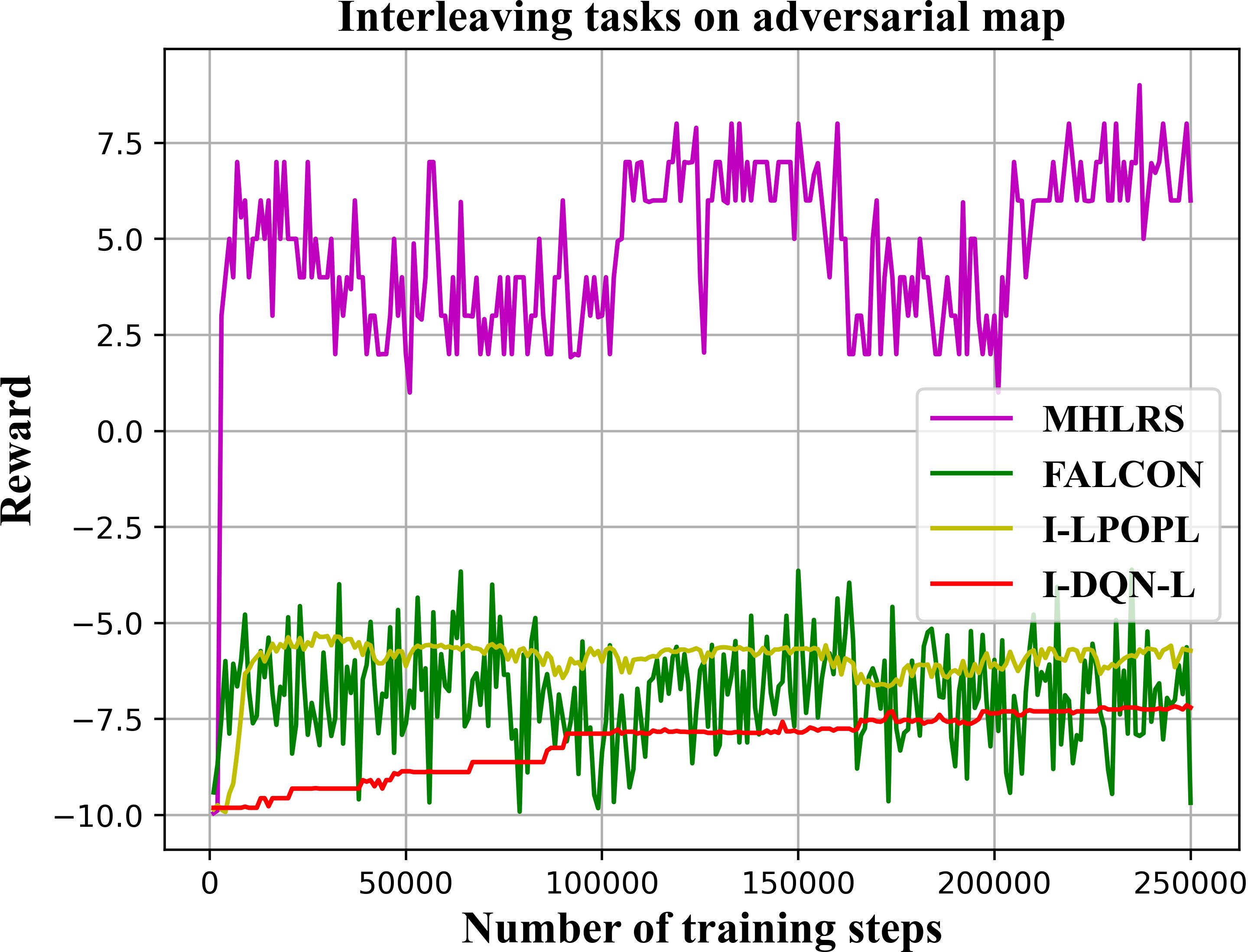}
 \caption{Reward curves in interleaving tasks (adversarial map).}
 \label{fig:Interleaving-adversarial}
 \end{figure}

 Figures \ref{fig:Interleaving-random}-\ref{fig:Interleaving-adversarial} and Table \ref{table3} show the results of the four algorithms tested on 10 tasks without unnecessary sequences. The results on FALCON are similar to Experiment 1. I-DQN-L still performs the worst, and MHLRS performs the best. MHLRS achieved a maximum reward of 9 on both types of maps, indicating that it has the potential for partial-ordered interleaving multi-task learning.

 \begin{figure}[htbp]
 \centering
 \includegraphics[height=4.5cm]{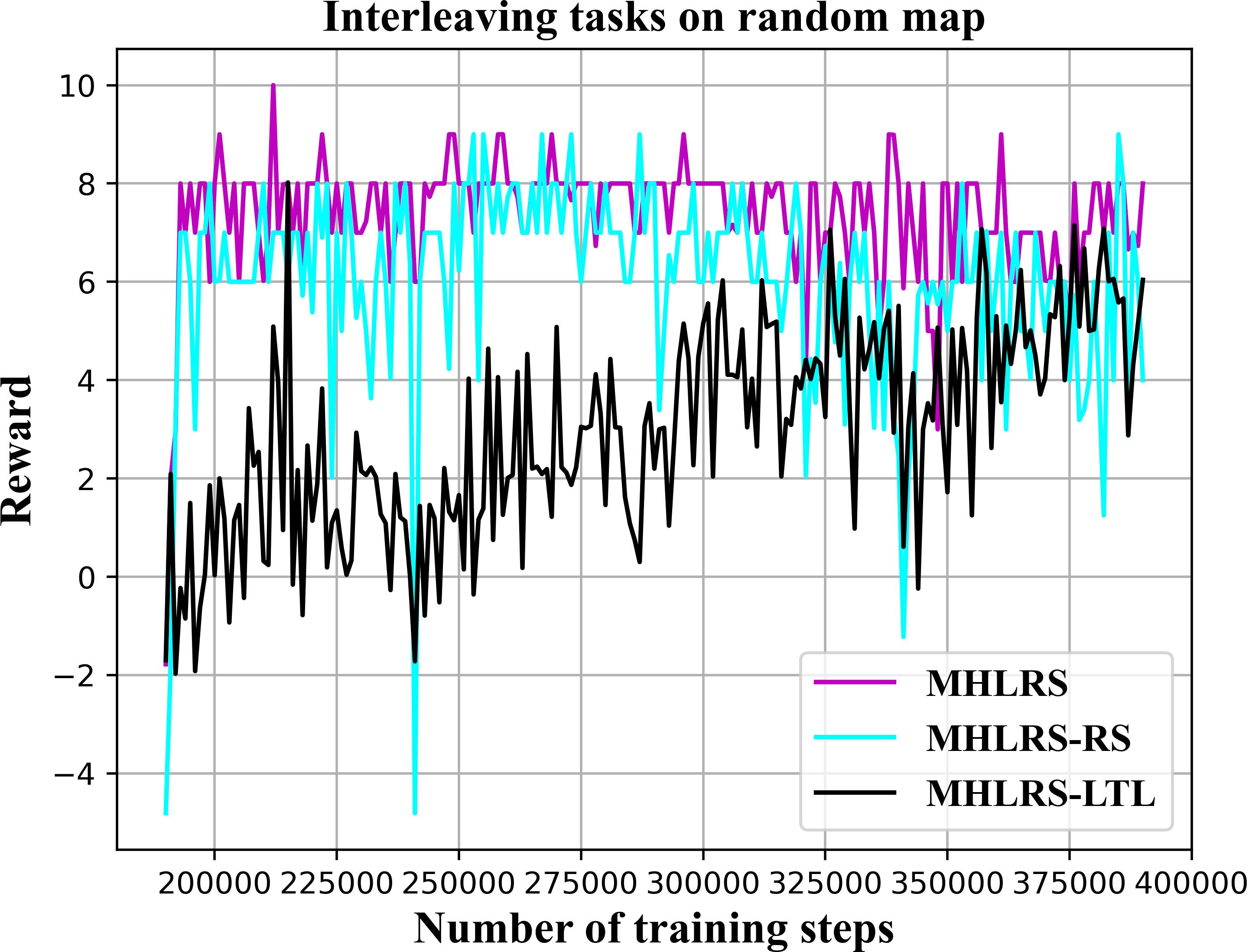}
 \caption{\vspace{-1ex}Reward curves of ablation experiment in interleaving tasks (random map).}
 \label{fig:Interleaving-random-ablation}
 \end{figure}

 \begin{figure}[htbp]
 \centering
 \includegraphics[height=4.5cm]{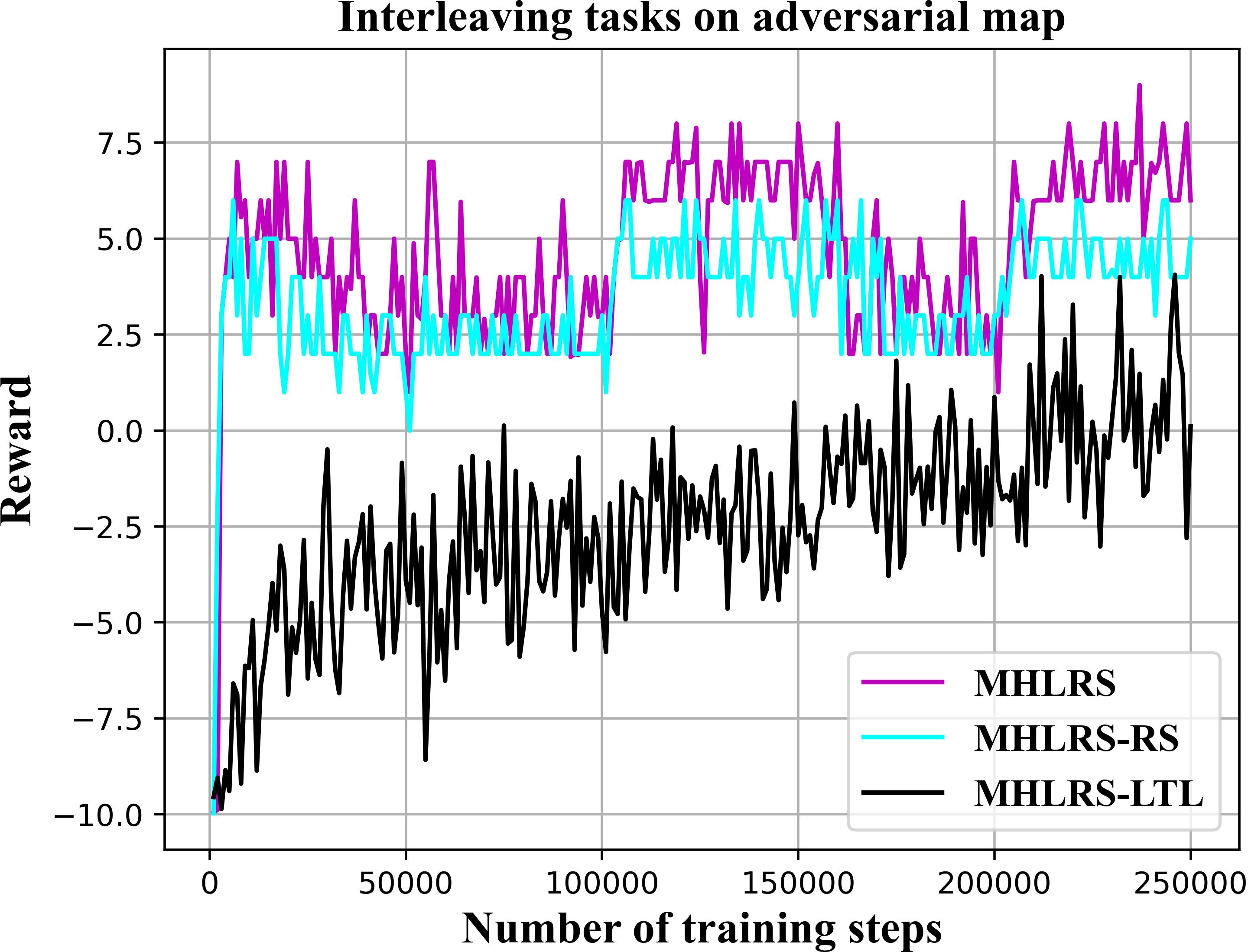}
 \caption{Reward curves of ablation experiment in interleaving tasks (adversarial map).}
 \label{fig:Interleaving-adversarial-ablation}
 \end{figure}
 
 \begin{table}[htbp]
\caption{Rewards of ablation experiment in the interleaving tasks.}\label{table2-ablation}
\centering
\resizebox{\linewidth}{8.5mm}{
\setlength{\tabcolsep}{1mm}
\begin{tabular}{c| c| c| c| c}
\hline
\multirow{2}{*}{Algorithm} & \multicolumn{2}{c|}{random map} & \multicolumn{2}{c}{adversarial map}\\
\cline{2-3}\cline{4-5}

& maximum reward & average reward& maximum reward & average reward\\
 
\hline
MHLRS	& \textbf{10.0} &	\textbf{4.71} & \textbf{9.0} &	\textbf{4.11}\\

MHLRS-RS 	& 	9.0 &	3.39 & 	6.0 &	2.89\\

MHLRS-LTL 	& 	8.02 & 	0.91  &	4.06 & 	-2.49\\

\hline
\end{tabular}}
\end{table}

Two separate experiments were conducted to demonstrate the effects of reward shaping and LTL. The results of these experiments are presented in Figures \ref{fig:Interleaving-random-ablation}-\ref{fig:Interleaving-adversarial-ablation} and Table \ref{table2-ablation}. The ablation experiment shows that in both types of maps, the maximum reward achieved by MHLRS-RS is slightly lower than MHLRS. Additionally, the algorithm becomes more unstable, and the average reward decreases by about 1.2 compared to MHLRS. Furthermore, the performance of MHLRS-LTL is worse than MHLRS, with both the average and maximum rewards being significantly lower on both random and adversarial maps. In fact, the average reward on the adversarial map is less than 0.

 \subsection{Experiment 3: Constrained Tasks}

The objective of the last experiment is to test the performance of these algorithms while ensuring safety. In this experiment, the agents are required to take shelter at night to avoid being attacked by zombies. To achieve this, a time limit is assigned to each of the 10 tasks. The time is measured as follows: every 10 steps represents one hour. The training begins at sunrise (5:00 am) and ends at sunset (9:00 pm).

For example, adding safety constraints to the task $``making\ $ $  fishing\_rod"$, $\varphi_{fishing\_rod}$ defined in Equation (\ref{equ13}) will be modified as Equation (\ref{equ15}):

 \begin{small}
\begin{equation}
\begin{aligned}
(is\_night \rightarrow at\_shelter)\ \mathcal{U} 
((\diamond(got\_wood 
\wedge \diamond(used\_toolshed \\ \wedge \diamond  used\_workbench))  
 \wedge \diamond  (got\_grass  \wedge \diamond  used\_workbench))\\
\wedge (is\_night \rightarrow at\_shelter)).
\end{aligned}
\label{equ15}
\end{equation}
 \end{small}

\noindent If the agent appears outside the shelter at night, we think the formula has been falsified, and the reward for the agent is -1.

\begin{table}[htbp]
\caption{Rewards in constrained tasks.}\label{table5}
\centering
\resizebox{\linewidth}{10mm}{
\setlength{\tabcolsep}{1mm}
\begin{tabular}{c| c| c| c| c}
\hline
\multirow{2}{*}{Algorithm} & \multicolumn{2}{c|}{random map} & \multicolumn{2}{c}{adversarial map}\\
\cline{2-3}\cline{4-5}

& maximum reward & average reward& maximum reward & average reward\\

\hline
MHLRS	& \textbf{1.8} &	\textbf{0.51}& \textbf{2.12} &	\textbf{0.35}\\

FALCON 	& 	0.71 &	-0.78 &	0.16 &	-0.96\\

I-LPOPL 	& 	-5.3 & 	-5.78 & -4.85 & 	-5.54\\

I-DQN-L 	& 	-5.89 &	-7.18	& 	-5.99 &	-7.09\\
\hline
\end{tabular}}
\end{table}

 \begin{figure}[htbp]
 \centering
 \includegraphics[height=4.5cm]{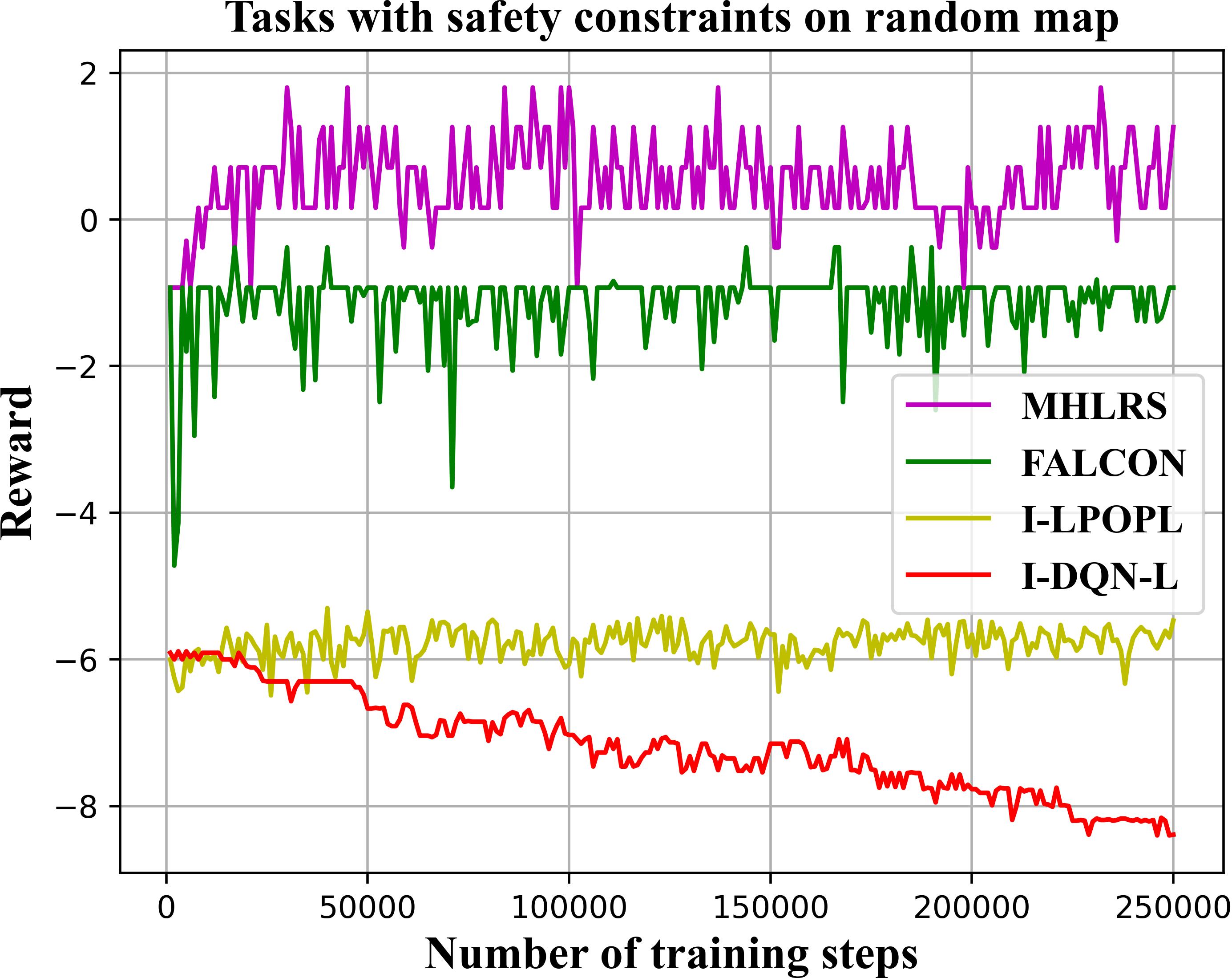}
 \caption{Reward curves in constrained tasks (random map).}
 \label{fig:Safety-random}
 \end{figure}

 \begin{figure}[htbp]
 \centering
 \includegraphics[height=4.5cm]{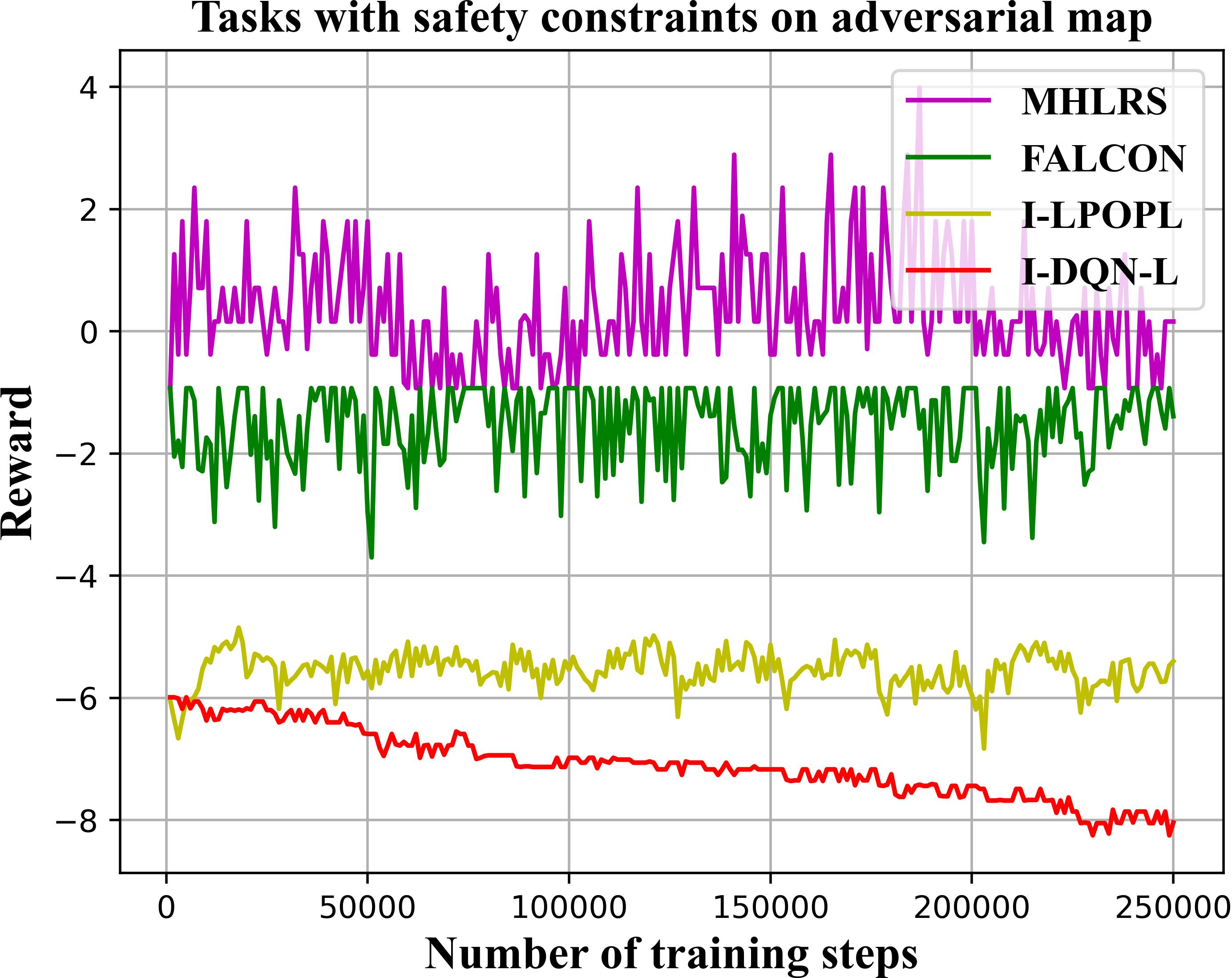}
 \caption{Reward curves in constrained tasks (adversarial).}
 \label{fig:Safety-adversarial}
 \end{figure}

The results of four algorithms for completing 10 tasks with security constraints are shown in Figures \ref{fig:Safety-random}-\ref{fig:Safety-adversarial} and Table \ref{table5}. The security constraints make it difficult to complete the tasks. I-DQN-L fails to learn effective strategies, and the rewards obtained on both types of maps are negative. I-LPOPL still does not get positive rewards, whose maximum reward is -4.85 (as can be seen in Table \ref{table5}). FALCON is superior to I-DQN-L and I-LPOPL, with average rewards slightly less than 0. MHLRS still performs the best, with the maximum reward close to 2 on both types of maps, showing its effectiveness in multi-agent learning with constraints.

 \begin{table}[htbp]
\caption{Rewards of ablation experiment in the constrained tasks.}\label{table3-ablation}
\centering
\resizebox{\linewidth}{8.5mm}{
\setlength{\tabcolsep}{1mm}
\begin{tabular}{c| c| c| c| c}
\hline
\multirow{2}{*}{Algorithm} & \multicolumn{2}{c|}{random map} & \multicolumn{2}{c}{adversarial map}\\
\cline{2-3}\cline{4-5}

& maximum reward & average reward& maximum reward & average reward\\
 
\hline
MHLRS	& \textbf{1.8} &	\textbf{0.51} & \textbf{3.99} &	\textbf{0.35}\\

MHLRS-RS 	& 	\textbf{1.8} &	0.29 & 	2.35 &	-0.20\\

MHLRS-LTL 	& 	0.71 & 	-0.58  &	0.16 & 	-0.83\\

\hline
\end{tabular}}
\end{table}

 \begin{figure}[htbp]
 \centering
 \includegraphics[height=4.5cm]{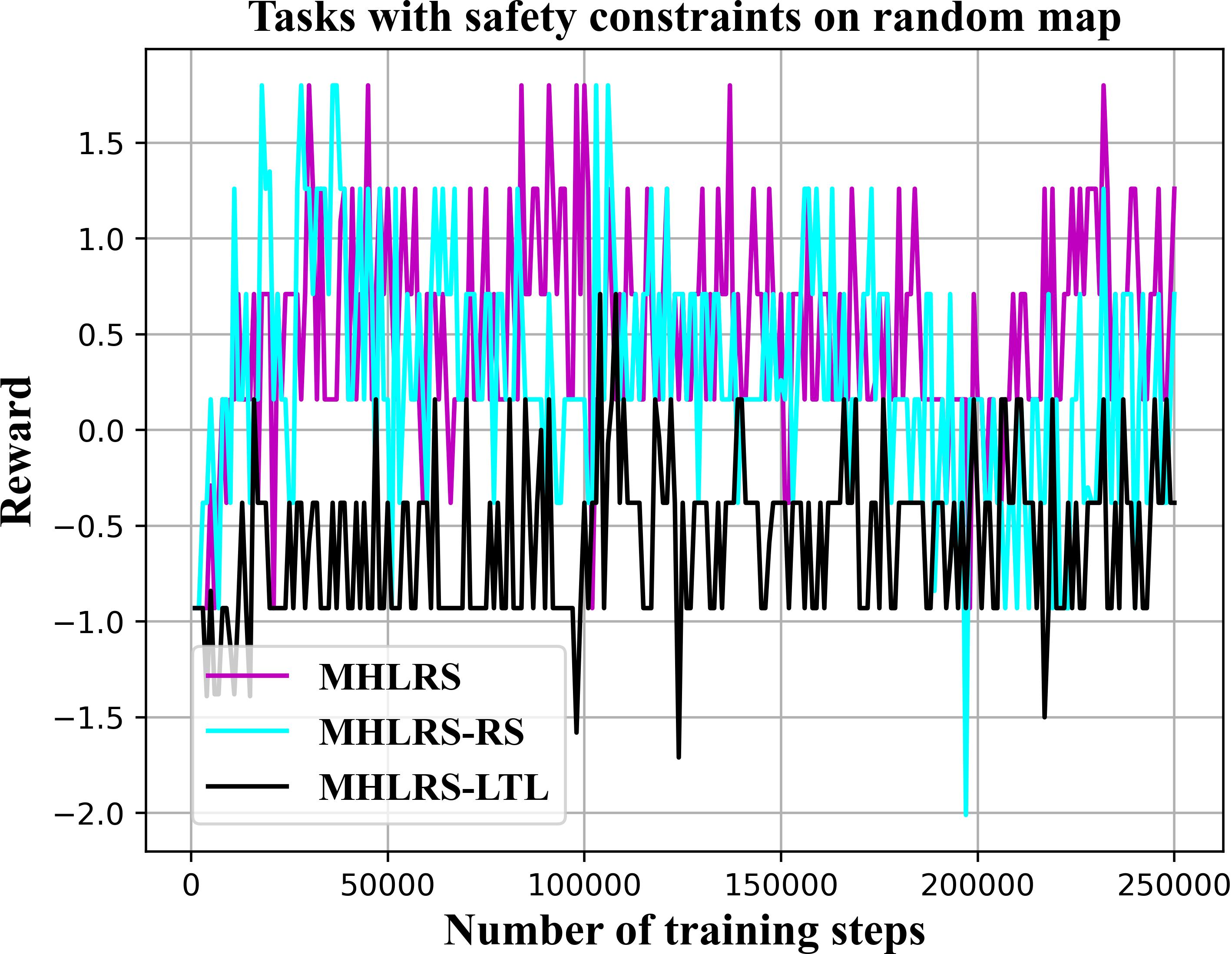}
 \caption{\vspace{-1ex}Reward curves of ablation experiment in constrained tasks (random map).}
 \label{fig:Safety-random-ablation}
 \end{figure}

\begin{figure}[htbp]
 \centering
 \includegraphics[height=4.5cm]{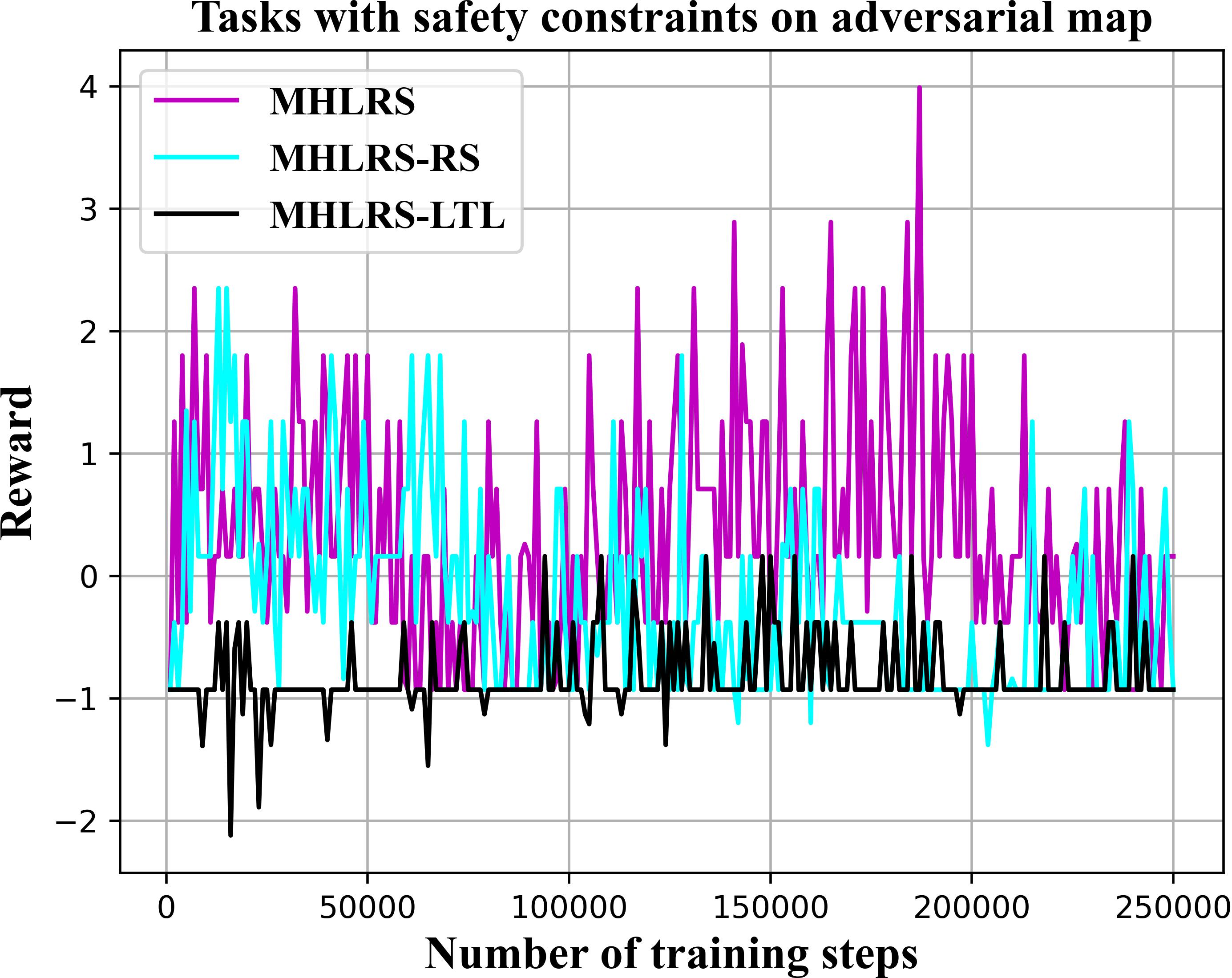}
 \caption{Reward curves of ablation experiment in constrained tasks (adversarial map).}
 \label{fig:Safety-adversarial-ablation}
 \end{figure}

We conducted ablation experiments on safety-constrained tasks, and the results are presented in Figures \ref{fig:Safety-random-ablation}-\ref{fig:Safety-adversarial-ablation} and Table \ref{table3-ablation}. According to the experiment results, MHLRS-RS achieved the same maximum reward as MHLRS on the random map. However, on the adversarial map, the maximum reward is much lower than MHLRS, and the average reward is also lower than MHLRS on both maps. On the other hand, the performance of MHLRS-LTL degraded significantly, with average and maximum rewards much lower than MHLRS on both maps. These ablation experiments proved that both reward shaping and LTL play important roles in our algorithm.

\section{Conclusion}

This work proposes a multi-agent hierarchical reinforcement learning algorithm, named MHLRS. In this algorithm, LTL is utilized to express the internal logic of multi-task and enhance the interpretability and credibility of agent decisions. Based on the techniques of value iteration and reward shaping, MHLRS can facilitate coordination and cooperation among multiple agents. The effectiveness of MHLRS has been demonstrated in experiments on sequence-based tasks, partial-ordered tasks, and safety-constrained tasks. Additionally, ablation experiments demonstrate the importance of reward shaping and LTL in the algorithm.

Future research directions include considering the integration of LTL with other logical forms, such as fuzzy logic \cite{10258120}, and dynamic logic \cite{Liu2016}, to expand the expressiveness of the task representation. We would also like to explore the incorporation of other off-policy reinforcement learning methods into the proposed framework in order to improve the adaptability and stability of the learning algorithms.

\bibliographystyle{plain}
\bibliography{refs}

\iffalse
\section{Biography Section}
\label{sec:Biography}
\begin{IEEEbiography}[{\vspace{-4ex}\includegraphics[width=0.9in,height=1.00in,clip,keepaspectratio]{author/chanjuan}}]{Chanjuan Liu}
received the PhD degree in computer software and theory from Peking University, Beijing, China, in 2016. She is currently an associate professor with School of Computer Science and Technology, Dalian University of Technology. Her current research interests include deep learning, modal logic.
\end{IEEEbiography}

\vspace{-4ex}

\begin{IEEEbiography}[{\vspace{-4ex}\includegraphics[width=0.9in,height=1.10in,clip,keepaspectratio]{author/cjm}}]{Jinmiao Cong}
received the B.E. degree in Computer Science and Technology from Dalian University of Technology. He is currently a master degree candidate in Computer Science and Technology from Dalian University of Technology. His research interests include deep reinforcement learning, multi-agent systems.
\end{IEEEbiography}

\vspace{-4ex}

\begin{IEEEbiography}[{\includegraphics[width=0.9in,height=1.20in,clip,keepaspectratio]{author/zeq}}]{Enqiang Zhu}
received the PhD degree in computer science from Peking University, China, in 2015. He is currently a professor with the Institute of Computing Science and Technology, Guangzhou University. His current research interests include control optimization.
\end{IEEEbiography}
 \fi

\vfill

\end{document}